\documentclass[11pt]{article}

\usepackage[final]{acl}

\usepackage{times}
\usepackage{latexsym}
\usepackage[T1]{fontenc}
\usepackage[utf8]{inputenc}
\usepackage{microtype}

\usepackage{graphicx}
\usepackage{booktabs}
\usepackage{array}
\usepackage{multirow}
\usepackage{subcaption}
\usepackage{float}
\usepackage{placeins}
\usepackage{dblfloatfix}  

\usepackage{amsmath,amssymb}
\usepackage{amsfonts}
\usepackage{nicefrac}
\usepackage{bm}

\usepackage{algorithm}
\usepackage[noend]{algpseudocode}
\algrenewcommand\algorithmicrequire{\textbf{Input:}}
\algrenewcommand\algorithmicensure{\textbf{Output:}}
\usepackage{enumitem}
\usepackage{cleveref}
\usepackage{pifont}

\newcommand{\eg}{\textit{e.g.}}

\newcommand{\cmark}{\ding{51}}
\newcommand{\xmark}{\ding{55}}

\hypersetup{
  pdftitle={Co-Evolutionary Prompt Optimization with Cross-Category Transfer for Zero-Shot Anomaly Detection},
  pdfauthor={Sisi Zhu, Changwei Yu, Renshuai Tao, Zhenliang Ni},
}

\title{Co-Evolutionary Prompt Optimization with Cross-Category Transfer\\
for Zero-Shot Anomaly Detection}

\author{
  Sisi Zhu\textsuperscript{1}\thanks{\ Equal contribution.},
  Changwei Yu\textsuperscript{1}\footnotemark[1],
  Renshuai Tao\textsuperscript{1}\thanks{\ Corresponding author.},
  Zhenliang Ni\textsuperscript{2}\footnotemark[2] \\
  \textsuperscript{1}Institute of Information Science, Beijing Jiaotong University, Beijing, China \\
  \textsuperscript{2}Institute of Automation, Chinese Academy of Sciences, Beijing, China \\
  \texttt{\{24281153, 24281148, rstao\}@bjtu.edu.cn} \\
  \texttt{nizhenliang17@mails.ucas.ac.cn} \\
}

\begin{document}

\maketitle

\begin{abstract}
Zero-shot anomaly detection (ZSAD) has gained significant attention for its practical value in industrial inspection. Recently, CLIP-based approaches have been widely adopted in ZSAD due to their strong vision-language generalization capabilities. However, existing methods commonly employ continuous prompt embeddings for prompt optimization and encode semantics in latent vectors, which lack interpretability and scalability. To this end, we propose CoEvoAD, a co-evolutionary framework for discrete prompt selection. CoEvoAD performs prompt search in the discrete natural-language space using an evolutionary algorithm. Candidate prompts are iteratively generated, evaluated, and selected throughout population evolution, thus preserving the interpretability and composability of natural language. Furthermore, we introduce a Cross-Category Transfer Objective (CCTO), which treats held-out source categories as proxies for unseen categories and scores prompt rules based on their estimated cross-category transferability, effectively improving cross-category generalization. Extensive experiments are conducted to validate the effectiveness of CoEvoAD, and the results show that it achieves state-of-the-art performance across multiple anomaly detection datasets. The code is available at \url{https://github.com/rstao-bjtu/CoEvoAD}.

\end{abstract}

\section{Introduction}
\label{sec:intro}

\begin{figure}[t]
  \centering
  \includegraphics[width=\columnwidth]{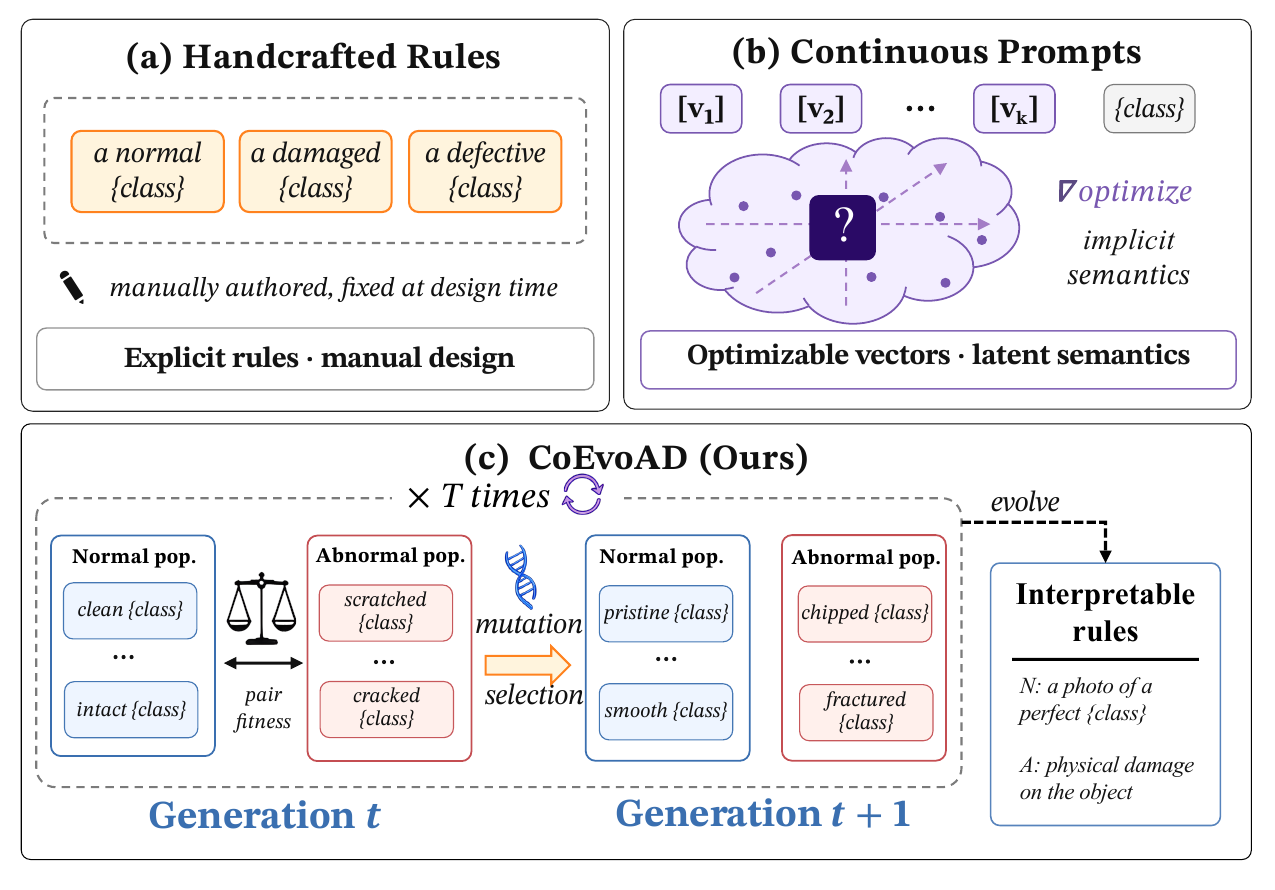}
  \caption{Prompt regimes for ZSAD: handcrafted templates, continuous prompts, and CoEvoAD.}
  \label{Fig:teaser}
\end{figure}

Industrial anomaly detection aims to identify defective samples and localize anomalous regions in visual inspection images. In practice, deploying a detector to a new product category often requires normal images from the target category and, in supervised settings, defect labels or pixel-level masks, followed by category-specific retraining or calibration. This requirement is especially costly when production lines change frequently and real defects are scarce. Built on CLIP~\cite{radford2021learning}, recent zero-shot anomaly detection (ZSAD) methods reduce this dependency by comparing visual features with normal and abnormal text prompts, enabling detection in unseen categories~\cite{jeong2023winclip,zhou2024anomalyclip,qu2025bayespfl}. In this setting, prompt selection becomes central to specifying transferable normal--abnormal rules. We study how interpretable prompt rules can be automatically generated and selected for cross-category transfer, as illustrated in \Cref{Fig:teaser}.

Despite recent progress in CLIP-based ZSAD, prompt construction and selection still face two limitations for cross-category transfer. First, existing prompt construction still struggles to jointly achieve prompt interpretability and scalability. Handcrafted rules remain explicit but fixed after design, while prompt-learning methods such as AnomalyCLIP~\cite{zhou2024anomalyclip}, AdaCLIP~\cite{cao2024adaclip}, and Bayes-PFL~\cite{qu2025bayespfl} improve adaptation by optimizing learnable prompt embeddings or distributions over them. Yet the learned prompt semantics are encoded in uninterpretable continuous representations rather than explicit natural-language rules, limiting prompt interpretability and rule-level reuse. Second, source-category performance alone does not provide reliable evidence of cross-category generalization. A prompt may perform well on the source categories used for selection yet fail on unseen categories, because its transferability across categories has not been evaluated. In summary, current prompt-selection methods in ZSAD still fail to achieve both interpretability and scalability. Moreover, candidate prompt selection relies solely on performance within the optimization categories, lacking cross-category validation signals.

To address the lack of prompt interpretability and scalability, we introduce CoEvoAD, a co-evolutionary discrete prompt-rule search framework. This framework performs search in the discrete natural language space using an evolutionary algorithm rather than relying on opaque black-box vector representations. Candidate prompt rules are continuously generated, evaluated, and selected throughout population evolution, while maintaining the interpretability, composability, and auditability of natural language, which provides strong potential for scalability. The fitness function combines the evaluation score on source data with the semantic contrast between normal and abnormal descriptions, thereby enabling a transparent and controllable optimization process without gradients. To avoid the objective conflict caused by mixing normal and abnormal rules in traditional methods, we maintain independent populations for the two roles and achieve co-evolution through a joint fitness design. This allows them to expand along their respective semantic directions in the large-scale natural language space while preserving the necessary semantic contrast. Overall, the framework transforms prompt-rule search from a single black-box vector into a structured, multi-role, multi-candidate co-evolutionary system, effectively improving interpretability and scalability.

To address the cross-category generalization issue, we introduce the Cross-Category Transfer Objective (CCTO), a prompt-rule selection criterion specifically designed for cross-category transfer. The key idea behind CCTO is to construct systematic held-out splits within the source domain, where a subset of source categories is treated as proxy unseen categories. This design enables the evaluation of candidate prompt rules under a simulated zero-shot transfer scenario, allowing us to assess their consistency and robustness when applied across categories. By leveraging only source-domain data, CCTO provides an evaluation signal that aligns more closely with the requirements of zero-shot generalization, without relying on any target-domain supervision. As a result, the proposed objective significantly improves the cross-category generalization ability of the selected rules, offering a more reliable selection criterion under the zero-shot setting.

\begin{figure*}[!t]
  \centering
  \includegraphics[width=\textwidth]{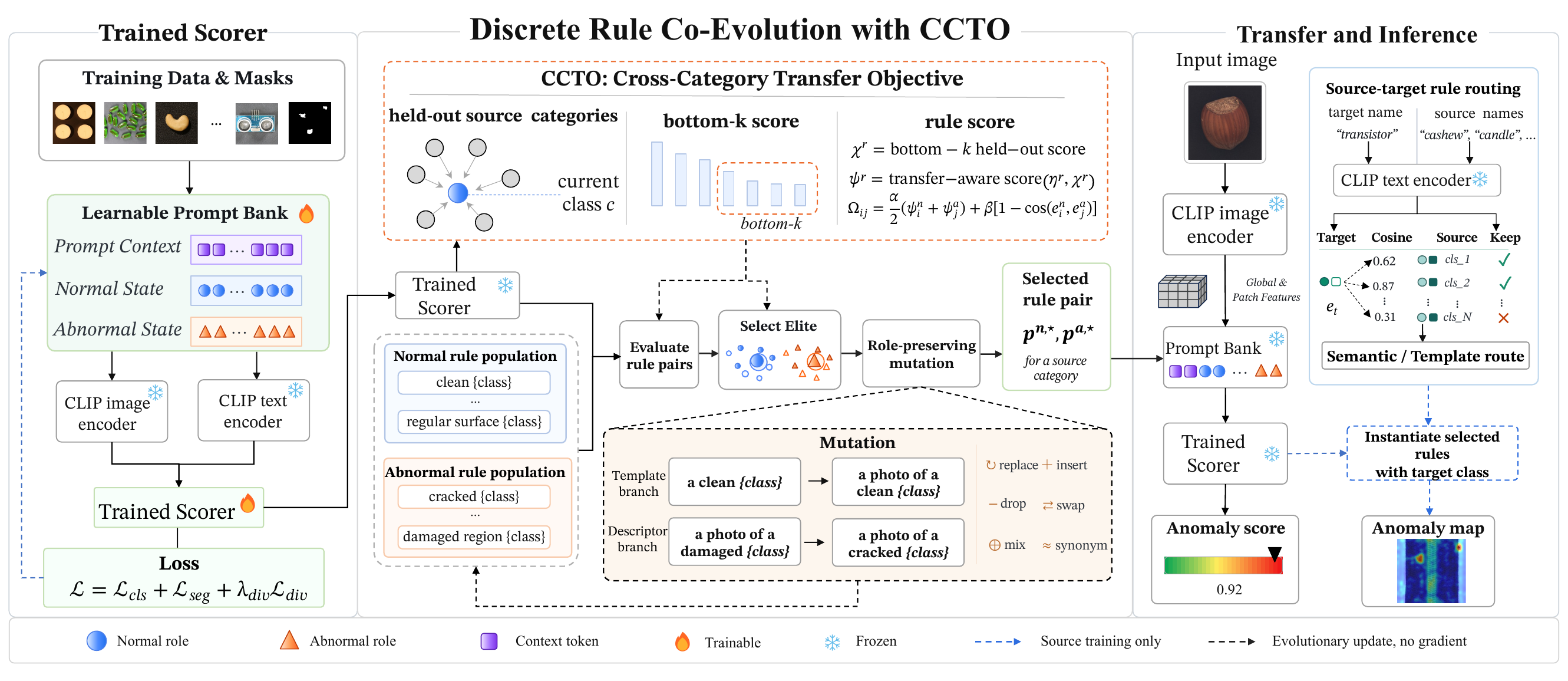}
  \caption{Overview of CoEvoAD. A frozen source-trained scorer evaluates role-separated normal and abnormal rule populations; CCTO uses held-out source categories to select rules for unseen-category inference.}
  \label{Fig:overview}
\end{figure*}

\section{Related Work}
\label{sec:related}

\paragraph{Traditional Anomaly Detection.}
Traditional anomaly detection usually learns category-specific normal patterns from target-category normal images. In feature space, PaDiM~\cite{defard2021padim} models normal statistics, while PatchCore~\cite{roth2022patchcore} stores patch-level memories. For discriminative training, DRAEM~\cite{zavrtanik2021draem} synthesizes defects, whereas SimpleNet~\cite{liu2023simplenet} perturbs features. Reverse Distillation~\cite{deng2022reverse} uses teacher--student discrepancies, with UniAD~\cite{you2022uniad} relying on reconstruction. However, these methods depend on target-category normal images, which limits deployment when new categories appear without prior normal data.

\paragraph{Zero-Shot Anomaly Detection.}
Zero-shot anomaly detection targets anomaly classification and localization on categories unseen during training. WinCLIP~\cite{jeong2023winclip} introduces hand-crafted normal/abnormal prompts for training-free anomaly detection. Subsequent methods shift from manual state words to optimized prompt representations. AnomalyCLIP~\cite{zhou2024anomalyclip} addresses category dependence with object-agnostic prompts, while AdaCLIP~\cite{cao2024adaclip} adopts static-dynamic prompt design. Bayes-PFL~\cite{qu2025bayespfl} further models prompt uncertainty through Bayesian prompt flow. Recent work broadens CLIP adaptation beyond text-side prompting. VCP-CLIP~\cite{qu2024vcpclip} uses visual context to guide prompt construction, whereas AA-CLIP~\cite{ma2025aaclip} strengthens anomaly-aware textual anchors. AF-CLIP~\cite{fang2025afclip} and AdaptCLIP~\cite{gao2026adaptclip} further adapt visual features or CLIP representations for anomaly detection. MRAD~\cite{xu2026mrad} introduces memory-driven retrieval. These methods improve CLIP-based anomaly reasoning through prompt learning, visual context, feature adaptation, or memory retrieval.

\paragraph{Prompt Tuning and Discrete Prompt Optimization.}
Prompt tuning reduces manual prompt engineering by optimizing prompt representations. CoOp~\cite{zhou2022coop} learns continuous context vectors for vision-language recognition. CoCoOp~\cite{zhou2022cocoop} makes the learned context condition-dependent. BBT~\cite{sun2022bbt} studies black-box prompt tuning without gradient access. Discrete prompt optimization keeps the search in token space. AutoPrompt~\cite{shin2020autoprompt} uses gradient-guided word substitution, RLPrompt~\cite{deng2022rlprompt} formulates prompt search as reinforcement learning, and EvoPrompt~\cite{guo2024evoprompt} applies population-based search to LLM prompts. CoEvoAD instead searches discrete normal and abnormal rule populations for ZSAD. Rule pairs are co-evolved with joint fitness and selected using cross-category evidence drawn from source-domain data.

\section{Method}
\label{sec:method}

\subsection{Overview}
\label{sec:overview}

CoEvoAD treats cross-category ZSAD as discrete prompt-rule selection, as shown in \Cref{Fig:overview}. The prompt-conditioned scorer is kept fixed during rule search, so the optimization only changes the natural-language rule strings used to render prompts. CoEvoAD maintains separate normal and abnormal populations, updates them through role-preserving mutation, and selects rule pairs with a joint fitness that combines evaluation scores and normal--abnormal semantic separation. CCTO further evaluates candidate rules on held-out categories from \(\mathcal{C}_s\), adding a transfer-oriented signal without target-domain feedback. At inference, selected rules are instantiated for unseen categories through a pre-specified transfer policy and scored by the same prompt-conditioned scorer to produce image-level anomaly scores and pixel-level anomaly maps; each rule remains interpretable through its template, role-specific descriptors, and class-name slot.

\subsection{Discrete Prompt-Rule Co-Evolution}
\label{sec:coevo_search}

CoEvoAD formulates prompt optimization as a discrete search over natural-language prompt rules. It uses the fixed prompt-conditioned scorer as the evaluator and varies only rule strings.

\paragraph{Role-typed rule representation and populations.}
CoEvoAD searches over structured natural-language rules rather than unconstrained prompt strings. Since anomaly scoring is defined over a normal--abnormal prompt pair, we organize the candidate space into two role-typed rule pools. For role $r \in \{\mathrm{n}, \mathrm{a}\}$ (normal and abnormal), a rule is represented as
\begin{equation}
\begin{aligned}
p^r &= (\tau^r, D^r, \langle\text{class}\rangle), \\
D^r &= (d^r_1,\ldots,d^r_m),
\end{aligned}
\label{eq:rule_tuple}
\end{equation}
where $\tau^r$ is a role-compatible template, $D^r$ is a sequence of descriptors drawn from the corresponding role vocabulary, and $\langle\text{class}\rangle$ is a class-name slot. The rendering function $z^r = R(p^r, c)$ fills the slot with category name $c$ and concatenates the template and descriptors into a CLIP text prompt. The normal role uses descriptors of acceptable appearance (\eg, \emph{clean}, \emph{intact}, \emph{regular}), while the abnormal role uses descriptors of defect evidence (\eg, \emph{damaged}, \emph{cracked}, \emph{contaminated}). This representation only defines the admissible candidate space; it does not manually determine the final prompt rule. Candidates are generated through role-preserving mutation and selected by the paired fitness function and CCTO, while keeping the role type fixed prevents normal and abnormal descriptions from being mixed during co-evolution.

We maintain two separate populations,
\begin{equation}
\begin{aligned}
\mathcal{P}^{\mathrm{n}} &= \{p^{\mathrm{n}}_1,\ldots,p^{\mathrm{n}}_N\}, \\
\mathcal{P}^{\mathrm{a}} &= \{p^{\mathrm{a}}_1,\ldots,p^{\mathrm{a}}_N\},
\end{aligned}
\label{eq:role_populations}
\end{equation}
where \(N\) is the population size and each \(p_i^r\) is a role-typed rule. Initial populations are constructed from role-specific templates and descriptor pools; since each candidate is an explicit natural-language rule, the same rule can be instantiated for another category by substituting the class-name placeholder.

Algorithm~\ref{alg:coevo} summarizes the rule search. The scorer is frozen during this stage; only the rendered natural-language rules are changed. Candidate rules are scored by CCTO-adjusted role scores and selected through pair-level fitness on rule pairs.

\paragraph{Candidate evaluation and pair fitness.}
For a fixed optimization category \(c\), candidates are scored on labeled source-domain evaluation data with the fixed prompt-conditioned scorer; the exact split is specified in \Cref{app:implementation}. Candidate pairs are formed with partners sampled from the opposite population.

Given a candidate pair \((p_i^{\mathrm{n}},p_j^{\mathrm{a}})\), the pair fitness is
\begin{equation}
\begin{aligned}
\Omega_{ij}
&=
\alpha \frac{\eta_i^{\mathrm{n}}+\eta_j^{\mathrm{a}}}{2}
+\beta \delta_{ij},\\
\delta_{ij}
&=
1-\cos(e_i^{\mathrm{n}},e_j^{\mathrm{a}}),
\end{aligned}
\label{eq:pair_fitness}
\end{equation}
where \(\eta_i^r\) is the resulting candidate score, \(e_i^r\) is the normalized text embedding of the instantiated rule \(R(p_i^r,c)\), and \(r \in \{\mathrm{n},\mathrm{a}\}\). The first term measures evaluation scores computed by the scorer, while the second keeps normal and abnormal descriptions separated in text space. Pairs with identical instantiated normal and abnormal strings are marked invalid and excluded from selection. When CCTO is enabled, \(\eta_i^r\) is replaced by the adjusted score \(\psi^r(p_i^r,c)\) defined in Section~\ref{sec:ccto}; the pair-fitness form is unchanged.

\begin{algorithm}[!t]
\small
\caption{Prompt-rule co-evolution with \mbox{CCTO selection}.}
\label{alg:coevo}
\begin{algorithmic}[1]
\Require Frozen scorer \(S_\theta\); source categories \(\mathcal{C}_s\); role-specific rule pools \(\mathcal{V}^{\mathrm n}, \mathcal{V}^{\mathrm a}\); population size \(N\), generations \(G\), elite size \(K_e\), partner budget \(K\)
\Ensure Rule cache \(\Pi=\{(c,\,p_c^{\mathrm{n}\star},\,p_c^{\mathrm{a}\star})\}_{c\in\mathcal{C}_s}\)

\For{$c \in \mathcal{C}_s$}
    \State \(\mathcal{H}_c \leftarrow \mathcal{C}_s \setminus \{c\}\) \Comment{held-out source categories}
    \State Initialize \(\mathcal{P}_c^{\mathrm n}\) and \(\mathcal{P}_c^{\mathrm a}\) from \(\mathcal{V}^{\mathrm n}\) and \(\mathcal{V}^{\mathrm a}\)
    \For{$g = 1,\ldots,G$}
        \For{$r \in \{\mathrm n,\mathrm a\}$}
            \State \(\psi^{r}(\cdot,c) \leftarrow \mathrm{CCTOScore}(\mathcal{P}_c^{r}, c, \mathcal{H}_c)\) \Comment{Eq.~\ref{eq:ccto_score}}
        \EndFor
        \State Sample a diverse subset of \(K\) abnormal partners
        \State Pair every normal rule with the subset; score pairs by \(\Omega\) \Comment{Eq.~\ref{eq:pair_fitness}}
        \State Rank candidates by mean fitness of their sampled pairs \Comment{unpaired: keep \(\psi^r\)}
        \State Keep the top \(K_e\) candidates as role-wise elites
        \State Refill \(\mathcal{P}_c^{\mathrm n}\) and \(\mathcal{P}_c^{\mathrm a}\) with role-preserving mutations
    \EndFor
    \State Re-score the top rules of each role on the full evaluation set
    \State \(\Pi[c] \leftarrow (\arg\max_{p^{\mathrm n}} \tilde{\eta}^{\mathrm n},\ \arg\max_{p^{\mathrm a}} \tilde{\eta}^{\mathrm a})\) \Comment{role-wise}
\EndFor
\State \Return \(\Pi\)
\end{algorithmic}
\end{algorithm}

\paragraph{Role-preserving mutation and selection.}
CoEvoAD updates the two populations through mutation only; no recombination across roles is used. Each new candidate is produced by mutating a same-role parent, and the mutation operates only on the mutable fields of the rule grammar in Eq.~\ref{eq:rule_tuple}. For descriptors, we apply replacement, insertion, deletion, local reordering, and synonym substitution within the same role-specific vocabulary. For templates, we sample from a small role-compatible template pool. All mutations preserve the role type of the parent rule, so a normal candidate remains a description of acceptable appearance and an abnormal candidate remains a description of visible defect evidence.

At each generation, a small diverse subset of abnormal candidates is sampled as shared partners, every normal candidate is paired with this subset, and each candidate is ranked by the average fitness of the sampled pairs it participates in; candidates that appear in no sampled pair retain their role-adjusted scores. Selection is performed separately within the normal and abnormal populations. We retain \(K_{\mathrm{elite}}\) high-fitness candidates in each role and fill the remaining slots with mutated candidates. After \(G\) generations, the top-ranked rules of each role are re-scored on the full source-side evaluation set, and the final normal and abnormal rules are selected independently by these re-evaluated role scores \(\tilde{\eta}^{r}\), yielding
\(\Pi_c=(p_c^{\mathrm{n}\star},p_c^{\mathrm{a}\star})\).

\paragraph{Rule transfer and inference.}
After search, CoEvoAD stores one selected normal--abnormal rule pair $\Pi_c=(p_c^{\mathrm{n}\star},p_c^{\mathrm{a}\star})$ for each source category $c\in\mathcal{C}_s$. For an unseen target category $c_t$, a pre-specified rule-transfer policy selects one stored pair using only category names and the saved source rules. The selected pair $(\tilde p_t^{\mathrm{n}},\tilde p_t^{\mathrm{a}})$ is rendered with $c_t$, yielding $z_t^r=R(\tilde p_t^r,c_t)$ for $r\in\{\mathrm{n},\mathrm{a}\}$. The fixed prompt-conditioned scorer then produces the image-level anomaly score and pixel-level anomaly map. The policy is fixed before target evaluation.

\subsection{Cross-Category Transfer Objective}
\label{sec:ccto}

Prompt rules selected only on the optimization category may capture category-specific visual patterns that do not transfer to unseen categories. CCTO turns prompt-rule selection into a category-held-out transfer test. Thus, candidate selection is guided by both within-category evaluation scores and cross-category consistency.

For a candidate rule \(p^r\) on category \(c\), CCTO defines the cross-category score as
\begin{equation}
\chi^r(p,c)
=
\frac{1}{\kappa}
\sum_{j=1}^{\kappa}\xi_{(j)},
\label{eq:ccto_cross}
\end{equation}
where \(\xi_{(1)} \le \cdots \le \xi_{(m)}\) are the held-out scores sorted in ascending order after evaluating \(p^r\) on each remaining source category, and \(\kappa=\min(k,m)\). We aggregate the held-out scores with a bottom-$k$ operator. A simple mean can be dominated by easy held-out categories and may overlook rules that fail on harder transfers. The bottom-$k$ score instead emphasizes the least transferable held-out categories while remaining less brittle than a strict minimum when the number of source categories is limited.

Normal and abnormal rules use the held-out scores in different ways. We define the role-adjusted selection scores as
\begin{equation}
\begin{aligned}
\psi^{\mathrm{n}}(p,c)
&=
(1-\lambda_{\mathrm{n}})\eta^{\mathrm{n}}
+\lambda_{\mathrm{n}}\chi^{\mathrm{n}},\\
\psi^{\mathrm{a}}(p,c)
&=
\eta^{\mathrm{a}}
+
\lambda_{\mathrm{a}}
\left(
\eta^{\mathrm{a}}
-
\chi^{\mathrm{a}}
\right),
\end{aligned}
\label{eq:ccto_score}
\end{equation}
where \(\eta^r=\eta^r(p,c)\) is the in-category score, \(\chi^r=\chi^r(p,c)\) is the cross-category score in Eq.~\ref{eq:ccto_cross}, and \(\lambda_{\mathrm{n}},\lambda_{\mathrm{a}}\) control the two role-specific adjustments. The normal branch interpolates between in-category performance and cross-category consistency. The abnormal branch uses the margin \(\eta^{\mathrm{a}}-\chi^{\mathrm{a}}\) to retain defect-discriminative abnormal rules and suppress overly generic anomaly descriptions. The adjusted score \(\psi^r(p,c)\) replaces \(\eta_i^r\) in Eq.~\ref{eq:pair_fitness}. All held-out categories are drawn from \(\mathcal{C}_s\).

\subsection{Overall Loss}
\label{sec:loss}

The prompt bank and scoring head are trained on the source dataset, while the CLIP image and text encoders remain frozen. For each source sample, the category name is used as the textual anchor; evolved prompt rules are introduced only during prompt-rule search. The scorer is optimized with an image-level binary classification loss, a pixel-level localization loss combining focal~\cite{lin2017focal} and Dice~\cite{milletari2016vnet} terms, and a same-role prompt-group diversity regularizer. We write the objective compactly as
\(\mathcal{L}_{\mathrm{train}}=\mathcal{L}_{\mathrm{cls}}+\mathcal{L}_{\mathrm{seg}}+\lambda_{\mathrm{div}}\mathcal{L}_{\mathrm{div}}\). In implementation, scorer training further includes mask-guided crop augmentation, inter-role margin regularization, and category-agnostic regularization. This training recipe is identical for the class-name control and all main CoEvoAD comparisons, so matched-control gains can be attributed to prompt-rule selection rather than scorer updates.

\begin{table*}[!t]
  \centering
  \caption{Image-level benchmark comparison (\%) across six anomaly detection datasets. Bold/underline denote best/second-best; ``--'' marks entries not reported by the original paper.}
  \label{tab:main-image}
  \footnotesize
  \setlength{\tabcolsep}{3.5pt}
  \renewcommand{\arraystretch}{1.05}
  \resizebox{\textwidth}{!}{%
    \begin{tabular}{l|cc|cc|cc|cc|cc|cc}
        \toprule
        \shortstack[l]{Method $\rightarrow$\\Dataset $\downarrow$}
        & \multicolumn{2}{c|}{\shortstack{WinCLIP\\(CVPR'23)}}
        & \multicolumn{2}{c|}{\shortstack{AnomalyCLIP\\(ICLR'24)}}
        & \multicolumn{2}{c|}{\shortstack{AdaCLIP\\(ECCV'24)}}
        & \multicolumn{2}{c|}{\shortstack{Bayes-PFL\\(CVPR'25)}}
        & \multicolumn{2}{c|}{\shortstack{MRAD\\(ICLR'26)}}
        & \multicolumn{2}{c}{\shortstack{CoEvoAD\\(Ours)}} \\
        \cmidrule(lr){2-13}
        & AUROC$\uparrow$ & AP$\uparrow$
        & AUROC$\uparrow$ & AP$\uparrow$
        & AUROC$\uparrow$ & AP$\uparrow$
        & AUROC$\uparrow$ & AP$\uparrow$
        & AUROC$\uparrow$ & AP$\uparrow$
        & AUROC$\uparrow$ & AP$\uparrow$ \\
        \midrule
        MVTec-AD
        & 91.8 & 95.1
        & 91.5 & 96.2
        & 92.0 & 96.4
        & 92.3 & 96.7
        & \textbf{94.0} & \textbf{97.4}
        & \underline{93.4} & \underline{96.8} \\

        VisA
        & 78.1 & 77.5
        & 82.1 & 85.4
        & 83.0 & 84.9
        & \underline{87.0} & \underline{89.2}
        & 85.7 & 88.3
        & \textbf{87.4} & \textbf{89.8} \\

        BTAD
        & 83.3 & 84.1
        & 89.1 & 91.1
        & 91.6 & 92.4
        & \underline{93.2} & \textbf{96.5}
        & 92.4 & 94.2
        & \textbf{94.4} & \underline{96.1} \\

        KSDD2
        & 93.5 & 77.9
        & 92.1 & 77.8
        & 95.9 & 95.9
        & \underline{97.3} & \underline{97.9}
        & 95.1 & 88.9
        & \textbf{97.4} & \textbf{97.9} \\

        DAGM
        & 89.6 & 90.4
        & 95.6 & 94.6
        & 96.5 & 95.7
        & 97.7 & 97.0
        & \textbf{98.4} & \textbf{98.6}
        & \underline{98.2} & \underline{97.6} \\

        RSDD
        & 85.3 & 65.3
        & 73.5 & 55.0
        & 89.1 & 70.8
        & \underline{94.1} & \underline{92.3}
        & -- & --
        & \textbf{98.9} & \textbf{98.9} \\

        \midrule
        Mean
        & 86.9 & 81.7
        & 87.3 & 83.4
        & 91.4 & 89.4
        & \underline{93.6} & \underline{94.9}
        & -- & --
        & \textbf{95.0} & \textbf{96.2} \\
        \bottomrule
      \end{tabular}%
  }
\end{table*}

\begin{table*}[!t]
  \centering
  \caption{Pixel-level benchmark comparison (\%; AUROC and Per-Region Overlap (PRO)). Bold/underline denote best/second-best across the six datasets; ``--'' marks entries not reported by the original paper.}
  \label{tab:main-pixel}
  \footnotesize
  \setlength{\tabcolsep}{3.5pt}
  \renewcommand{\arraystretch}{1.05}
  \resizebox{\textwidth}{!}{%
    \begin{tabular}{l|cc|cc|cc|cc|cc|cc}
        \toprule
        \shortstack[l]{Method $\rightarrow$\\Dataset $\downarrow$}
        & \multicolumn{2}{c|}{\shortstack{WinCLIP\\(CVPR'23)}}
        & \multicolumn{2}{c|}{\shortstack{AnomalyCLIP\\(ICLR'24)}}
        & \multicolumn{2}{c|}{\shortstack{AdaCLIP\\(ECCV'24)}}
        & \multicolumn{2}{c|}{\shortstack{Bayes-PFL\\(CVPR'25)}}
        & \multicolumn{2}{c|}{\shortstack{MRAD\\(ICLR'26)}}
        & \multicolumn{2}{c}{\shortstack{CoEvoAD\\(Ours)}} \\
        \cmidrule(lr){2-13}
        & AUROC$\uparrow$ & PRO$\uparrow$
        & AUROC$\uparrow$ & PRO$\uparrow$
        & AUROC$\uparrow$ & PRO$\uparrow$
        & AUROC$\uparrow$ & PRO$\uparrow$
        & AUROC$\uparrow$ & PRO$\uparrow$
        & AUROC$\uparrow$ & PRO$\uparrow$ \\
        \midrule
        MVTec-AD
        & 85.1 & 64.6
        & 91.1 & 81.4
        & 86.8 & 33.8
        & 91.8 & \underline{87.4}
        & \textbf{93.0} & 86.8
        & \underline{92.2} & \textbf{87.9} \\

        VisA
        & 79.6 & 56.8
        & 95.5 & 87.0
        & 95.1 & 71.3
        & 95.6 & \underline{88.9}
        & \textbf{95.9} & 88.0
        & \underline{95.8} & \textbf{89.4} \\

        BTAD
        & 71.4 & 32.8
        & 93.3 & 69.3
        & 87.7 & 17.1
        & 93.9 & \underline{76.6}
        & \textbf{95.4} & 72.8
        & \underline{94.7} & \textbf{81.9} \\

        KSDD2
        & 97.9 & 91.2
        & 99.1 & 85.6
        & \underline{99.4} & 92.7
        & 96.1 & 70.8
        & 98.9 & \underline{95.6}
        & \textbf{99.6} & \textbf{98.5} \\

        DAGM
        & 83.2 & 55.4
        & 99.1 & 93.6
        & 97.0 & 40.9
        & \underline{99.3} & \underline{98.0}
        & 97.4 & 90.3
        & \textbf{99.5} & \textbf{98.3} \\

        RSDD
        & 95.1 & 75.4
        & 99.1 & 92.0
        & 99.5 & 50.5
        & \underline{99.6} & \underline{98.0}
        & -- & --
        & \textbf{99.8} & \textbf{99.0} \\

        \midrule
        Mean
        & 85.4 & 62.7
        & \underline{96.2} & 84.8
        & 94.3 & 51.1
        & 96.1 & \underline{86.6}
        & -- & --
        & \textbf{96.9} & \textbf{92.5} \\
        \bottomrule
      \end{tabular}%
  }
\end{table*}

\section{Experiments}
\label{sec:experiments}

\subsection{Experimental Setup}
\label{sec:experimental_setup}

\paragraph{Datasets.}
We evaluate CoEvoAD on six industrial anomaly detection benchmarks: MVTec-AD~\cite{bergmann2019mvtec} (15 categories of manufactured objects), VisA~\cite{zou2022spot} (12 categories of complex inspection objects), BTAD~\cite{mishra2021vtadl} (three industrial product categories), KSDD2~\cite{bozic2021mixed} (industrial steel surface defects), DAGM~\cite{dagm2007} (manually re-annotated synthetic-texture defects), and RSDD~\cite{niu2021unsupervised} (rail surface defects).

\paragraph{Protocol.}
Following prior CLIP-based ZSAD settings~\cite{zhou2024anomalyclip,qu2025bayespfl}, CoEvoAD is evaluated under strict cross-dataset transfer: the scorer is trained on a source dataset and evaluated on target datasets without target-domain images, labels, or supervision. The two primary transfer directions are VisA$\to$MVTec-AD and MVTec-AD$\to$VisA. The remaining datasets (BTAD, KSDD2, DAGM, RSDD) serve as external industrial targets evaluated under the same source-only protocol.

\paragraph{Baselines.}
We compare against representative CLIP-based ZSAD baselines: WinCLIP~\cite{jeong2023winclip}, AnomalyCLIP~\cite{zhou2024anomalyclip}, AdaCLIP~\cite{cao2024adaclip}, Bayes-PFL~\cite{qu2025bayespfl}, and MRAD~\cite{xu2026mrad}. Baseline numbers in \Cref{tab:main-image,tab:main-pixel} are taken from the original papers or their official codebases where available.

\paragraph{Evaluation metrics.}
Following common ZSAD evaluation, we report image-level AUROC and AP and pixel-level AUROC and Per-Region Overlap (PRO) for benchmark comparison.
For matched controls and ablations, we additionally report pixel-level AP and F1, averaged over categories.

\paragraph{Implementation details.}
CoEvoAD uses frozen CLIP ViT-L/14@336px~\cite{radford2021learning}, with images resized to $518\times518$.
The prompt-bank scorer is trained for 30 epochs and frozen before prompt-rule search.
Unless otherwise stated, we use $M=3$ prompt groups, context length $L_c=5$, state length $L_s=5$, population size $N=16$, $G=5$ search generations, $K_e=4$ elites, and $K=3$ sampled partners.

\subsection{Main Results}
\label{sec:main_results}

We compare CoEvoAD against representative CLIP-based ZSAD baselines (\Cref{tab:main-image,tab:main-pixel}) and against a matched class-name prompt control under the same scorer and protocol (\Cref{tab:coevoad_vs_control}). The two comparisons cover cross-paper benchmark context and within-paper attribution, respectively.

\begin{table}[!t]
  \centering
  \caption{Matched comparison. Both systems use the same frozen source-trained scorer and protocol; the control uses class-name prompts without rule search. \(\Delta = \mathrm{CoEvoAD} - \mathrm{Control}\). Seed 111; three-seed means and standard deviations for the two primary directions are reported in \Cref{tab:multiseed}.}
  \label{tab:coevoad_vs_control}
  \footnotesize
  \setlength{\tabcolsep}{4pt}
  \renewcommand{\arraystretch}{1.0}
  \resizebox{\columnwidth}{!}{%
  \begin{tabular}{llrrr}
  \toprule
  Target & Metric & Control & CoEvoAD & $\Delta$ \\
  \midrule
  \multirow{3}{*}{\textbf{MVTec-AD}}
    & Image AUROC $\uparrow$ & 93.01 & 93.39 & \textbf{+0.38} \\
    & Pixel AP $\uparrow$    & 47.48 & 47.72 & \textbf{+0.24} \\
    & Pixel F1 $\uparrow$    & 47.69 & 47.96 & \textbf{+0.27} \\
  \midrule
  \multirow{3}{*}{\textbf{VisA}}
    & Image AUROC $\uparrow$ & 87.05 & 87.43 & \textbf{+0.38} \\
    & Pixel AP $\uparrow$    & 30.51 & 31.11 & \textbf{+0.60} \\
    & Pixel F1 $\uparrow$    & 36.21 & 36.72 & \textbf{+0.51} \\
  \midrule
  \multirow{3}{*}{\textbf{BTAD}}
    & Image AUROC $\uparrow$ & 92.81 & 92.68 & $-0.13$        \\
    & Pixel AP $\uparrow$    & 41.05 & 41.66 & \textbf{+0.61} \\
    & Pixel F1 $\uparrow$    & 44.69 & 46.25 & \textbf{+1.56} \\
  \midrule
  \multirow{3}{*}{\textbf{RSDD}}
    & Image AUROC $\uparrow$ & 97.69 & 97.62 & $-0.07$        \\
    & Pixel AP $\uparrow$    & 46.39 & 49.30 & \textbf{+2.91} \\
    & Pixel F1 $\uparrow$    & 49.11 & 50.10 & \textbf{+0.99} \\
  \bottomrule
  \end{tabular}}
\end{table}

\begin{figure*}[!t]
  \centering
  \includegraphics[width=\textwidth]{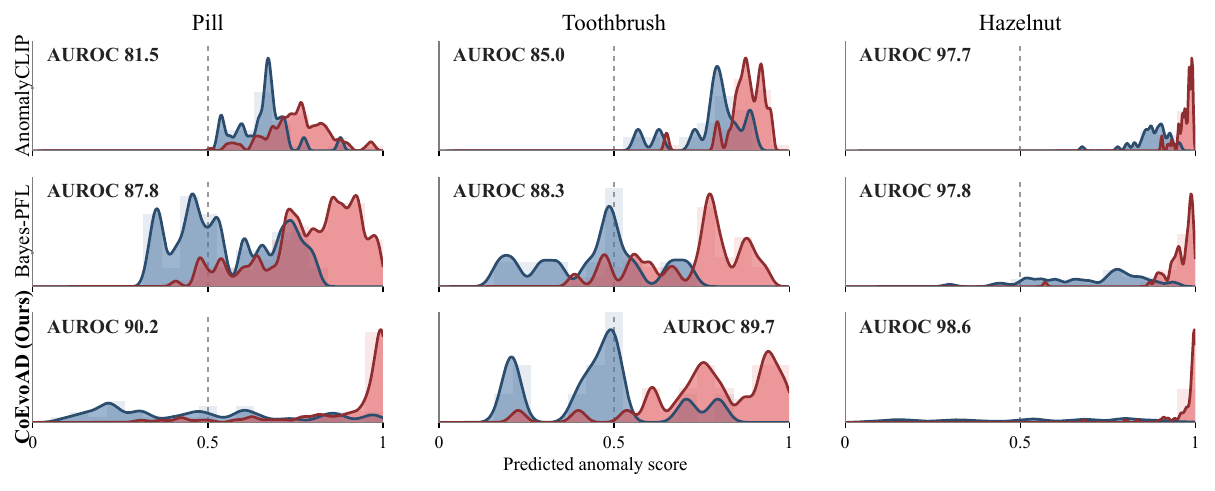}
  \caption{Per-image anomaly score distributions on three representative MVTec-AD categories (Pill, Toothbrush, Hazelnut) under the VisA$\to$MVTec-AD matched protocol. Rows compare AnomalyCLIP, Bayes-PFL, and CoEvoAD; blue/red denote normal/abnormal samples; per-cell AUROC is shown.}
  \label{Fig:score_distribution_v2m}
\end{figure*}

CoEvoAD separates normal and anomalous samples more clearly than the baselines on representative MVTec-AD categories (\Cref{Fig:score_distribution_v2m}).

CoEvoAD attains the highest mean image AUROC/AP (95.0/96.2) and pixel AUROC/PRO (96.9/92.5) among the compared CLIP-based ZSAD baselines (\Cref{tab:main-image,tab:main-pixel}), with the largest image gain on RSDD.
\Cref{tab:coevoad_vs_control} reports the matched-protocol comparison: CoEvoAD improves the class-name control on both primary transfer directions and yields consistent pixel-level AP and F1 gains on the external industrial targets under the matched MVTec-AD-source setting.

\paragraph{Robustness across categories.}
\Cref{tab:bottomk-delta} tests whether the matched-control pixel gains are driven by high-baseline categories alone. Bottom-$k$ $\Delta$ ($k{=}3$) exceeds the mean $\Delta$ in all four rows, indicating that the average lift is not concentrated on easy high-baseline categories.

\begin{table}[!t]
  \centering
  \caption{Per-category $\Delta$ statistics (mean, bottom-$k$, worst) vs the matched control.}
  \label{tab:bottomk-delta}
  \footnotesize
  \setlength{\tabcolsep}{4pt}
  \renewcommand{\arraystretch}{1.05}
  \begin{tabular}{l l r r r}
    \toprule
    Dir. & Metric & Mean $\Delta$ & Bot-$k$ $\Delta$ & Worst $\Delta$ \\
    \midrule
    \multirow{2}{*}{MVTec} & Pixel AP$\uparrow$ & +0.24 & \textbf{+0.51} & $-$0.48 \\
                            & Pixel F1$\uparrow$ & +0.27 & \textbf{+0.45} & $-$0.41 \\
    \midrule
    \multirow{2}{*}{VisA}  & Pixel AP$\uparrow$ & +0.60 & \textbf{+0.78} & $-$3.27 \\
                            & Pixel F1$\uparrow$ & +0.51 & \textbf{+0.77} & $-$3.37 \\
    \bottomrule
  \end{tabular}
\end{table}

\begin{figure*}[!t]
  \centering
  \includegraphics[width=0.96\textwidth]{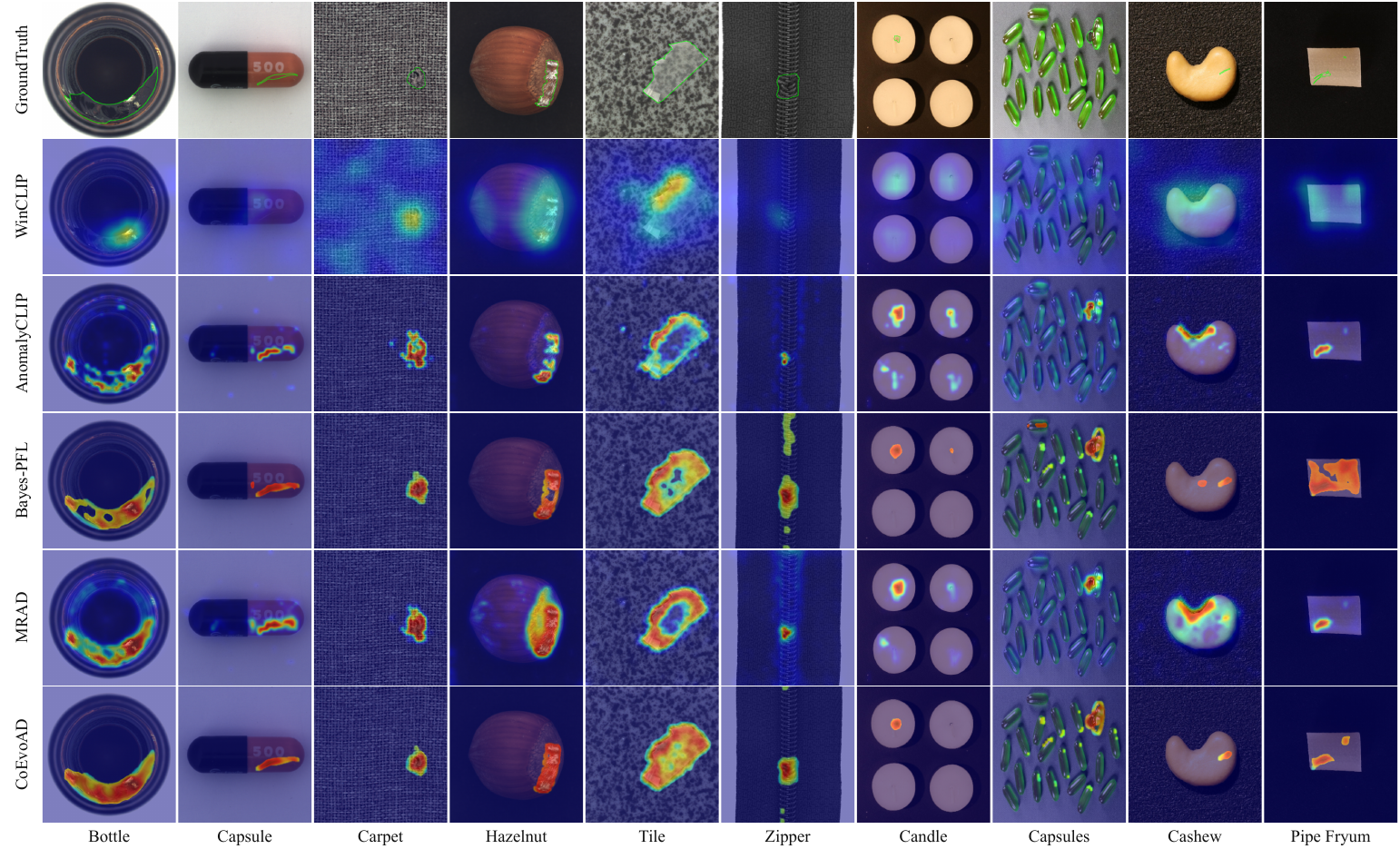}
  \caption{Qualitative localization examples on MVTec-AD and VisA. Columns show representative target categories; rows compare ground-truth contours and anomaly maps from CLIP-based baselines and CoEvoAD. Green contours denote annotated anomaly regions; warmer colors indicate higher anomaly scores.}
  \label{Fig:qualitative_maps}
\end{figure*}

\begin{table}[!t]
  \centering
  \caption{Same-budget search controls. R0 is the class-name reference; search rows report percentage-point changes over R0.}
  \label{tab:same_budget_search}
  \footnotesize
  \setlength{\tabcolsep}{2.5pt}
  \renewcommand{\arraystretch}{1.05}
  \makebox[\columnwidth][r]{\resizebox{\columnwidth}{!}{%
  \begin{tabular}{@{}lrrrr@{}}
    \toprule
    & \multicolumn{2}{c}{\textbf{MVTec-AD}} & \multicolumn{2}{c}{\textbf{VisA}} \\
    \cmidrule(lr){2-3}\cmidrule(lr){4-5}
    Method & \shortstack{Image\\AUROC $\uparrow$} & \shortstack{Pixel\\AP $\uparrow$} & \shortstack{Image\\AUROC $\uparrow$} & \shortstack{Pixel\\AP $\uparrow$} \\
    \midrule
    R0 class-name & 93.01 & 47.48 & 87.05 & 30.51 \\
    \addlinespace[3pt]
    Random search & +0.08 & $-$0.01 & $-$0.95 & +0.52 \\
    Single-pop. EA & +0.14 & $-$0.05 & $-$0.24 & +0.07 \\
    GA crossover & +0.17 & $-$0.01 & $-$0.23 & \textbf{+0.74} \\
    Role-sep. CoEvo & +0.05 & +0.04 & $-$0.03 & +0.47 \\
    \addlinespace[3pt]
    \textbf{CoEvoAD} & \textbf{+0.38} & \textbf{+0.24} & \textbf{+0.38} & +0.60 \\
    \bottomrule
  \end{tabular}
  }}
\end{table}

\paragraph{Same-budget search controls.}
CoEvoAD denotes role-separated co-evolution with CCTO. Random search and single-population variants do not consistently improve both transfer directions, suggesting that the gain is not explained by candidate budget alone. CoEvoAD gives the strongest overall balance across Image AUROC and Pixel AP, although the crossover control obtains the highest Pixel AP on MVTec-AD$\to$VisA.

CoEvoAD produces responses more concentrated around annotated defect regions on representative MVTec-AD and VisA categories (\Cref{Fig:qualitative_maps}).

\subsection{Rule Inspection}
\label{sec:rule_inspection}

\begin{table}[!t]
  \centering
  \caption{Representative natural-language rules selected by CoEvoAD (the leading learned soft-context token is omitted for compactness).}
  \label{tab:selected-rules}
  \fontsize{9}{13}\selectfont
  \setlength{\tabcolsep}{3pt}
  \begin{tabular}{l >{\raggedright\arraybackslash}p{0.32\columnwidth} >{\raggedright\arraybackslash}p{0.41\columnwidth}}
    \toprule
    Category & Normal rule & Abnormal rule \\
    \midrule
    screw & clean surface screw & broken shape with missing parts \\
    bottle & normal bottle & irregular visible surface bottle \\
    grid & regular standard grid & faulty flawed abnormal grid \\
    toothbrush & a photo of a healthy toothbrush & faulty toothbrush \\
    pcb2 & flawless fine pcb2 & structural defect like a hole or cut \\
    cashew & a photo of healthy cashew & faulty flawed cashew \\
    candle & clean intact candle & visible anomaly on object \\
    \bottomrule
  \end{tabular}
\end{table}

\looseness=-1
Normal rules retain explicit category anchors, while abnormal rules combine category-specific defect cues with general defect descriptors (\Cref{tab:selected-rules}), consistent with CCTO's role-aware selection. For instance, abnormal rules name concrete defect modes for structured objects (\eg, ``structural defect like a hole or cut'' for \texttt{pcb2}; ``broken shape with missing parts'' for \texttt{screw}), while less structured categories (\eg, \texttt{candle}) retain generic descriptors like ``visible anomaly on object''. The selected rules are interpretable text and can be edited, audited, and instantiated with another category name through the class-name slot, the same mechanism used by the rule-transfer policy (Section~\ref{sec:coevo_search}).

\subsection{Ablation Studies}
\label{sec:ablation}

\begin{table}[!t]
  \centering
  \caption{Component ablation comparing matched class-name prompt control, role-separated co-evolution (CoEvo), and CoEvoAD (CoEvo $+$ CCTO). Bold marks the best score per row. Seed 111 (\Cref{tab:multiseed} reports multi-seed variation).}
  \label{tab:ablation}
  \footnotesize
  \setlength{\tabcolsep}{2.5pt}
  \renewcommand{\arraystretch}{0.98}
  \begin{tabular}{@{}llrrr@{}}
    \toprule
    Target & Metric & Baseline & CoEvo & \shortstack{CoEvo\\+CCTO} \\
    \midrule
    \multirow{3}{*}{\textbf{MVTec-AD}}
      & Image AUROC$\uparrow$ & 93.01 & 93.06 & \textbf{93.39} \\
      & Pixel AP$\uparrow$    & 47.48 & 47.52 & \textbf{47.72} \\
      & Pixel F1$\uparrow$    & 47.69 & 47.67 & \textbf{47.96} \\
    \midrule
    \multirow{3}{*}{\textbf{VisA}}
      & Image AUROC$\uparrow$ & 87.05 & 87.02 & \textbf{87.43} \\
      & Pixel AP$\uparrow$    & 30.51 & 30.98 & \textbf{31.11} \\
      & Pixel F1$\uparrow$    & 36.21 & 36.33 & \textbf{36.72} \\
    \midrule
    \multirow{3}{*}{\textbf{BTAD}}
      & Image AUROC$\uparrow$ & 92.81 & \textbf{92.87} & 92.68 \\
      & Pixel AP$\uparrow$    & 41.05 & 40.85 & \textbf{41.66} \\
      & Pixel F1$\uparrow$    & 44.69 & 45.22 & \textbf{46.25} \\
    \midrule
    \multirow{3}{*}{\textbf{RSDD}}
      & Image AUROC$\uparrow$ & 97.69 & \textbf{97.92} & 97.62 \\
      & Pixel AP$\uparrow$    & 46.39 & 43.37 & \textbf{49.30} \\
      & Pixel F1$\uparrow$    & 49.11 & 46.69 & \textbf{50.10} \\
    \bottomrule
  \end{tabular}
\end{table}

\paragraph{Component analysis.}
Role-separated co-evolution alone yields mixed effects on image-level AUROC, suggesting role separation is necessary but not sufficient. Adding CCTO produces consistent positive gains on pixel-level localization (Pixel AP and Pixel F1 improve over the class-name control across all four targets), with weaker image-level effects. This supports our claim that held-out source categories serve as a more transfer-oriented selection signal than performance on the optimization categories. In the same-budget search controls (\Cref{tab:same_budget_search}), the non-role-separated baselines (Random / EA / GA) and role-separated co-evolution alone each fall short of the CoEvoAD pairing, which reaches $+0.38$ image AUROC under the matched protocol, indicating that the gain comes from the substrate--selection pairing rather than either component in isolation. This pattern reflects CCTO's role as a selection criterion rather than an additive scoring term.

\FloatBarrier
\paragraph{Inference latency.}
\Cref{tab:app-latency} compares per-image inference latency under one implementation-level protocol: single RTX~4090, batch size~1, 30 warm-up followed by 300 timed images per direction, with the timer wrapping the entire per-image call. CoEvoAD runs the same frozen scorer call as the class-name control, differing only in the prompt string: the two differ by under $0.7$\,ms with opposite signs across the two directions, so no systematic overhead is attributable to the evolved rules; the routing lookup, measured separately, adds $0.19/0.27$\,ms per image (under $0.5\%$). CoEvoAD's inference is $1.4$--$4.0\times$ faster than AnomalyCLIP, MRAD, AdaCLIP, and Bayes-PFL re-timed under the same protocol. Training costs are not comparable across released pipelines, so the one-time search cost (\Cref{tab:app-searchcost}) is reported separately; in the matched-control setting the scorer training is identical.

\begin{table}[!t]
  \centering
  \caption{Per-image inference latency (ms, mean over 300 timed images), timed on the VisA$\to$MVTec-AD direction under one protocol. $\dagger$: WinCLIP runs its native backbone and resolution, disclosed for completeness but not directly comparable.}
  \label{tab:app-latency}
  \scriptsize
  \setlength{\tabcolsep}{3pt}
  \renewcommand{\arraystretch}{1.1}
  \resizebox{\columnwidth}{!}{%
  \begin{tabular}{l l r}
    \toprule
    Method & Backbone / res. & ms/img \\
    \midrule
    Bayes-PFL (reproduced) & ViT-L/14@336px / 518 & 221.0 \\
    AdaCLIP (released ckpt) & ViT-L/14@336px / 518 & 114.56 \\
    MRAD (released ckpt) & ViT-L/14@336px / 518 & 103.55 \\
    AnomalyCLIP (reproduced) & ViT-L/14@336px / 518 & 75.97 \\
    WinCLIP$^{\dagger}$ (training-free) & ViT-B/16+240 / 240 & 56.08 \\
    Class-name control & ViT-L/14@336px / 518 & 55.00 \\
    \textbf{CoEvoAD} & ViT-L/14@336px / 518 & \textbf{55.64} \\
    \bottomrule
  \end{tabular}}
\end{table}

\section{Conclusion}
\label{sec:conclusion}

We presented CoEvoAD, a framework that searches discrete, interpretable normal and abnormal prompt rules in natural language and uses a Cross-Category Transfer Objective to favor rules generalizing across unseen target categories. Across six industrial anomaly detection benchmarks, CoEvoAD attains the highest mean image-level and pixel-level scores among compared CLIP-based ZSAD baselines, yielding positive matched-control gains attributable to prompt-rule search. The framework shows that interpretable prompt search is feasible for ZSAD via role-separated co-evolution guided by held-out cross-category signals. The saved rules remain auditable as natural-language strings, complementing continuous prompt embeddings.

\section*{Limitations}
\label{sec:limitations}

The benefits of CoEvoAD concentrate on pixel-level localization rather than image-level AUROC: matched-control gains in image AUROC are modest and not uniformly positive, and CCTO gains vary across target datasets. The gains are not uniform across target categories: the MVTec-AD$\to$VisA direction shows a worst-category pixel-AP regression on \texttt{pcb4} (\Cref{tab:app-route-audit-m2v}). CoEvoAD also inherits the limitations of the underlying CLIP backbone for defects that are not well described by natural language.

\section*{Ethics Statement}
\label{sec:ethics}

This work targets zero-shot anomaly detection for industrial visual inspection. We use public benchmark datasets and do not collect human-subject data or personally identifiable information. The method is intended to assist defect screening rather than replace human inspection in safety-critical production. Since anomaly detectors may produce false positives or false negatives under distribution shift, practical deployment should require human oversight, site-specific validation, and continuous monitoring.

Although CoEvoAD searches interpretable natural-language prompt rules, the selected rules may still reflect biases of the source datasets and the pretrained vision-language model. We therefore recommend auditing selected rules before deployment and avoiding deployment in high-stakes settings without further validation.

\section*{Acknowledgments}
This work was supported by the National Natural Science Foundation of China (No. 62506030), and the Beijing Natural Science Foundation (No. L242021), and the State Key Laboratory of Advanced Rail Autonomous Operation (Project No.~RAO 2026K06), Beijing Jiaotong University.

ChatGPT (OpenAI) was used solely for language polishing and limited code-debugging assistance. It was not used to generate scientific ideas, experimental designs, results, analyses, claims, or bibliographic references. All AI-assisted outputs were independently reviewed and verified by the authors.

\FloatBarrier
\bibliography{references}
\clearpage

\appendix
\setcounter{topnumber}{6}
\setcounter{bottomnumber}{6}
\setcounter{totalnumber}{12}
\setcounter{dbltopnumber}{6}
\renewcommand{\topfraction}{0.99}
\renewcommand{\bottomfraction}{0.99}
\renewcommand{\textfraction}{0.01}
\renewcommand{\floatpagefraction}{0.5}
\renewcommand{\dbltopfraction}{0.99}
\renewcommand{\dblfloatpagefraction}{0.5}
\makeatletter
\setlength{\@fptop}{0pt}
\setlength{\@fpsep}{10pt plus 4pt}
\setlength{\@fpbot}{0pt plus 1fil}
\setlength{\@dblfptop}{0pt}
\setlength{\@dblfpsep}{12pt plus 4pt}
\setlength{\@dblfpbot}{0pt plus 1fil}
\makeatother
\renewcommand{\thefigure}{A\arabic{figure}}
\setcounter{figure}{0}

\section{Experimental Protocol and Dataset Details}
\label{app:protocol}

\paragraph{Datasets.}
We evaluate CoEvoAD on six industrial anomaly detection benchmarks. MVTec-AD~\cite{bergmann2019mvtec} contains 15 object/texture categories with pixel-level defect masks; VisA~\cite{zou2022spot} provides 12 categories of small electronics and food items; BTAD~\cite{mishra2021vtadl} covers three industrial product categories; KSDD2~\cite{bozic2021mixed} and RSDD~\cite{niu2021unsupervised} focus on surface-defect inspection on metal and rail-surface, respectively; DAGM~\cite{dagm2007} consists of synthetic texture patches. Two additional texture-oriented external targets, DTD-Synthetic and a 7-class DAGM subset, are reported only as diagnostic transfers (Section~\ref{app:full_results}). All datasets are publicly available. Test sets follow the ZSAD dataset construction of Bayes-PFL~\cite{qu2025bayespfl}: MVTec-AD, VisA, BTAD, and DTD-Synthetic keep their standard test splits, while KSDD2 and RSDD are rebuilt from the official splits into balanced test sets. \Cref{tab:app-dataset-stats} reports the per-dataset image counts as loaded by our evaluation pipeline.

\begin{table}[H]
  \centering
  \caption{Test-set statistics of the evaluation benchmarks. KSDD2 and RSDD are rebuilt from their official splits into balanced test sets following Bayes-PFL~\cite{qu2025bayespfl}; the DAGM row refers to the 7-class subset used in our protocol (\Cref{tab:texture_external}).}
  \label{tab:app-dataset-stats}
  \small
  \resizebox{\columnwidth}{!}{%
  \begin{tabular}{l r r r}
    \toprule
    Dataset & Classes & Normal & Anomalous \\
    \midrule
    MVTec-AD~\cite{bergmann2019mvtec} & 15 & 467 & 1,258 \\
    VisA~\cite{zou2022spot} & 12 & 962 & 1,200 \\
    BTAD~\cite{mishra2021vtadl} & 3 & 451 & 290 \\
    KSDD2~\cite{bozic2021mixed} & 1 & 356 & 356 \\
    RSDD~\cite{niu2021unsupervised} & 1 & 387 & 387 \\
    DAGM (7-class subset)~\cite{dagm2007} & 7 & 1,350 & 1,350 \\
    DTD-Synthetic & 12 & 357 & 947 \\
    \bottomrule
  \end{tabular}}
\end{table}

\paragraph{Cross-dataset zero-shot setting.}
We evaluate two primary transfer directions: VisA$\to$MVTec-AD and MVTec-AD$\to$VisA. We refer to these directions by their target dataset name, namely \emph{MVTec} (for VisA$\to$MVTec-AD) and \emph{VisA} (for MVTec-AD$\to$VisA), when context is unambiguous. For external industrial targets (BTAD, KSDD2, RSDD, DAGM, DTD-Synthetic), we use two source conventions tied to the reporting purpose. The VisA-source convention applies a VisA-trained checkpoint to non-VisA targets and is used in the headline comparison, matching prior CLIP-based ZSAD baselines~\cite{zhou2024anomalyclip,qu2025bayespfl}. The MVTec-AD-source convention applies an MVTec-AD-trained checkpoint to non-MVTec-AD targets and is used in the matched-ablation tables, so that the control and CoEvoAD share a single frozen scorer with the MVTec-AD$\to$VisA primary direction. For the VisA target, an MVTec-AD-trained checkpoint is used in both settings. Source category labels are used during source-side prompt-bank training and prompt-rule search; target category names are used only as a frozen source-only transfer-policy key. No target-domain image, label, mask, statistic, or metric ever enters training, validation, calibration, prompt search, or model selection. Target data are observed only at final inference and final evaluation.

\section{Implementation Details}
\label{app:implementation}

\Cref{tab:app-hparams} lists the full configuration used for every CoEvoAD row in the main paper, shared across both transfer directions unless stated otherwise.

\paragraph{Source-side evaluation split for rule search.}
During prompt-rule search, candidate rules are evaluated on the labeled test partition of the \emph{source} dataset, used as a source-side validation pool (\texttt{--stage2\_split test} in the released code). The composite candidate score weights image AUROC, pixel AP, and pixel F1 as $0.4/0.3/0.3$. Search-time scoring uses 5 own-category evaluation batches and 20 CCTO batches; the final per-role re-scoring pass uses the full cached evaluation set. This partition belongs entirely to the source domain: no target-domain image, label, mask, statistic, or metric is used before final evaluation.

\begin{table}[!t]
  \centering
  \caption{Full implementation and hyperparameters.}
  \label{tab:app-hparams}
  \scriptsize
  \setlength{\tabcolsep}{4pt}
  \begin{tabular}{@{}p{0.46\columnwidth}p{0.50\columnwidth}@{}}
    \toprule
    Component & Setting \\
    \midrule
    Backbone & CLIP ViT-L/14@336px \\
    Input resolution & $518\times518$ \\
    Stage~1 epochs / optimiser & 30 / AdamW, cosine annealing \\
    Stage~1 learning rates & prompt $1{\times}10^{-3}$, other $1{\times}10^{-4}$ \\
    Prompt groups $M$ / context $L_c$ / state $L_s$ & 3 / 5 / 5 \\
    Search population $N$ & 16 \\
    Search generations $G$ & 5 \\
    Search elite count $K_e$ & 4 \\
    Search pairing count $K$ & 3 \\
    Rule-search evaluation split & source test partition \\
    Rule-search score weights (img AUROC / px AP / px F1) & 0.4 / 0.3 / 0.3 \\
    Own-category / CCTO evaluation batches & 5 / 20 \\
    Final per-role re-scoring & top 8, full eval set \\
    Stage~1 / rule-search batch size & 32 / 2 \\
    CCTO bottom-$k$ & 3 \\
    CCTO shared-role coefficient $\alpha_{\mathrm{ccto}}$ & 0.6 \\
    CCTO role-aware coefficients $\lambda_n$ / $\lambda_a$ & 0.35 / 0.20 \\
    Fitness weights $\alpha$ / $\beta$ & 0.85 / 0.15 \\
    Routing (MVTec / VisA) & semantic fallback / template transfer \\
    Semantic-fallback template, min-sim & \texttt{a photo of \{\}}, 0.45 \\
    Seeds & 111, 222, 333 \\
    Hardware & single NVIDIA RTX~4090 \\
    \bottomrule
  \end{tabular}
\end{table}

\paragraph{Stage~1 training design.}
\label{app:source-refinements}
The prompt-conditioned scorer applies three training components on top of the base classification + segmentation losses. \textit{Mask-guided crop augmentation} samples crops biased toward annotated defect regions during training, ensuring the pixel-level head sees fine-grained defect patterns. \textit{Inter-role margin regularization} adds a hinge-style term that pushes apart the per-image average embeddings of the normal and abnormal prompt groups, preventing the two roles from collapsing into a shared direction. \textit{Category-agnostic regularization} encourages the prompt-group representations to be transferable across categories by penalizing per-category specialization on the source set. Their isolated effects on VisA$\to$MVTec-AD are quantified in \Cref{tab:stage1}.

\begin{table}[!t]
  \centering
  \caption{Ablation of Stage~1 training components on the class-name control. Each row removes one component from the fixed scorer training.}
  \label{tab:stage1}
  \scriptsize
  \setlength{\tabcolsep}{2.5pt}
  \begin{tabular}{l rr}
    \toprule
    Variant & AUROC & AP\textsubscript{px} \\
    \midrule
    Full training setup (control) & 93.01 & 47.48 \\
    $-$ mask-guided crop & 92.94\,{\scriptsize($-$0.07)} & 46.35\,{\scriptsize($-$1.13)} \\
    $-$ inter-role margin & 92.83\,{\scriptsize($-$0.18)} & 47.04\,{\scriptsize($-$0.44)} \\
    $-$ agnostic reg. & 93.01\,{\scriptsize($\pm$0.00)} & 47.09\,{\scriptsize($-$0.39)} \\
    \bottomrule
  \end{tabular}
\end{table}

\section{Prompt-Rule Search Details}
\label{app:search_details}

\paragraph{Search algorithm.}
The rule search follows Algorithm~\ref{alg:coevo}, applied independently per source category with the hyperparameters in~\Cref{tab:app-hparams}.

\paragraph{Search cost.}
\Cref{tab:app-searchcost} reports the rule-search budget and observed wall-clock. The budget is quoted as \emph{per-role} scoring slots on the search split, $N \times G \times |\mathcal{C}_{\mathrm{src}}|$. Both populations are scored once per generation and once more after the last generation, so the search split incurs $2N(G{+}1)|\mathcal{C}_{\mathrm{src}}|$ slots. A separate final re-ranking pass then re-scores the top $K_r{=}8$ candidates of each role on the full evaluation set, adding $2K_r|\mathcal{C}_{\mathrm{src}}|$ evaluations. Pair fitness reuses cached role-branch scores and adds no image inference. The search is per source category and scales with the source category count. Wall-clock is the observed search span from the log of one representative seed on a single RTX~4090.

\begin{table}[!t]
  \centering
  \caption{Rule-search cost. $|\mathcal{C}_{\mathrm{src}}|$: number of source categories. Evals/role $= N{\times}G{\times}|\mathcal{C}_{\mathrm{src}}|$; both roles are scored each generation.}
  \label{tab:app-searchcost}
  \scriptsize
  \setlength{\tabcolsep}{4pt}
  \begin{tabular}{l c c c c r r}
    \toprule
    Dir. & $|\mathcal{C}_{\mathrm{src}}|$ & $N$ & $G$ & $K$ & Evals/role & Wall-clock \\
    \midrule
    MVTec (src VisA) & 12 & 16 & 5 & 3 & 960 & $\sim$3\,h\,47\,m \\
    VisA (src MVTec-AD) & 15 & 16 & 5 & 3 & 1200 & $\sim$5\,h\,34\,m \\
    \bottomrule
  \end{tabular}
\end{table}

\paragraph{Scaling with source categories.}
\Cref{tab:app-scaling} reports a measured sweep over the source-category count under the VisA$\to$MVTec-AD search configuration. The number of per-role candidate scoring slots grows exactly linearly ($N{\times}G{\times}|\mathcal{C}_{\mathrm{src}}|$), while the cost per slot grows sublinearly with $|\mathcal{C}_{\mathrm{src}}|$ because the held-out renderings are embedding-cached; total wall-clock is therefore superlinear but below naive quadratic. Extrapolating the per-evaluation trend to $|\mathcal{C}_{\mathrm{src}}|{=}15$ gives ${\approx}16$--$17$\,s/eval, matching the independent MVTec-AD$\to$VisA run in \Cref{tab:app-searchcost} ($16.7$\,s/eval). The search is a one-time offline cost with no LLM calls in the loop and parallelizes across source categories.

\begin{table}[!t]
  \centering
  \caption{Measured search-cost scaling with the source-category count (VisA$\to$MVTec-AD configuration, single RTX~4090). Evals/role $=N{\times}G{\times}|\mathcal{C}_{\mathrm{src}}|$; s/eval is wall-clock divided by this per-role count.}
  \label{tab:app-scaling}
  \scriptsize
  \setlength{\tabcolsep}{5pt}
  \begin{tabular}{c r r r}
    \toprule
    $|\mathcal{C}_{\mathrm{src}}|$ & Evals/role & Wall-clock (h) & s/eval \\
    \midrule
    3  & 240 & 0.34 & 5.1 \\
    6  & 480 & 1.12 & 8.4 \\
    9  & 720 & 2.38 & 11.9 \\
    12 & 960 & 3.71 & 13.9 \\
    \bottomrule
  \end{tabular}
\end{table}

\paragraph{Candidate-pool provenance.}
The frozen candidate pools used for the selection diagnostic in \Cref{app:ccto_details} were regenerated under the paper's exact search configuration (identical fitness definition, CCTO weights, and scorer checkpoint); the original search runs predate the pool-tracing tooling. Pool regeneration precedes any target evaluation and feeds nothing back into selection.

\paragraph{Same-budget controls with CCTO-augmented variants.}
\Cref{tab:app-same-budget-full} extends the main paper same-budget control table (\Cref{tab:same_budget_search}) with three CCTO-augmented variants that pair CCTO selection with non-role-separated search backbones (random prompt-pair search, single-population evolutionary algorithm, and genetic algorithm with crossover). Adding CCTO on top of these non-role-separated backbones does not consistently improve over the corresponding backbone-only rows, indicating that CCTO's selection benefit is coupled to the role-separated co-evolution substrate rather than functioning as a generic add-on. Values are absolute scores; the budget and frozen-scorer setting are identical to the main paper version.

\begin{table*}[!t]
  \centering
  \caption{Extended same-budget prompt-search control table with CCTO-augmented variants. All search rows use the same frozen scorer and candidate budget; R0 is the non-search class-name reference; seed~111.}
  \label{tab:app-same-budget-full}
  \footnotesize
  \setlength{\tabcolsep}{3pt}
  \renewcommand{\arraystretch}{1.05}
  \begin{tabular}{l c c rr rr}
    \toprule
    & & & \multicolumn{2}{c}{\textbf{MVTec-AD}} & \multicolumn{2}{c}{\textbf{VisA}} \\
    \cmidrule(lr){4-5}\cmidrule(lr){6-7}
    Method & Role-sep. & CCTO & AUROC\textsubscript{img} & AP\textsubscript{px} & AUROC\textsubscript{img} & AP\textsubscript{px} \\
    \midrule
    Class-name control (R0)                 & \xmark & \xmark & 93.01 & 47.48 & 87.05 & 30.51 \\
    \midrule
    Random prompt-pair search               & \xmark & \xmark & 93.09 & 47.47 & 86.10 & 31.03 \\
    Random prompt-pair search $+$ CCTO      & \xmark & \cmark & 93.23 & 47.45 & 86.90 & 30.80 \\
    Single-population evolutionary algorithm           & \xmark & \xmark & 93.15 & 47.43 & 86.81 & 30.58 \\
    Single-population evolutionary algorithm $+$ CCTO  & \xmark & \cmark & 93.24 & 47.54 & 86.90 & 30.62 \\
    Genetic algorithm with crossover            & \xmark & \xmark & 93.18 & 47.47 & 86.82 & \textbf{31.25} \\
    Genetic algorithm with crossover $+$ CCTO   & \xmark & \cmark & 93.19 & 47.45 & 86.95 & 30.35 \\
    Role-separated co-evolution             & \cmark & \xmark & 93.06 & 47.52 & 87.02 & 30.98 \\
    \midrule
    CoEvoAD (CoEvo$+$CCTO)                  & \cmark & \cmark & \textbf{93.39} & \textbf{47.72} & \textbf{87.43} & 31.11 \\
    \bottomrule
  \end{tabular}
\end{table*}

\section{Cross-Category Transfer Objective Details}
\label{app:ccto_details}

\paragraph{Bottom-$k$ aggregation.}
Let $\mathcal{C}_{\mathrm{src}}$ be the source category set used during prompt-bank training, and let $c \in \mathcal{C}_{\mathrm{src}}$ be the category currently being optimized. The held-out source set is $\mathcal{C}_{\mathrm{src}} \setminus \{c\}$. For a candidate pair $(p^{n}, p^{a})$, we compute its in-category fitness $\eta^{r}(p,c)$ on $c$ and its cross-category fitness $\chi^{r}(p, c')$ on each held-out $c' \in \mathcal{C}_{\mathrm{src}} \setminus \{c\}$. We aggregate per-$c'$ scores with the bottom-$k$ operator described in Sec.~\ref{sec:ccto}, using $k{=}3$ given the small source-category counts ($|\mathcal{C}_{\mathrm{src}}|=12$ for VisA$\to$MVTec-AD, $15$ for MVTec-AD$\to$VisA).

\paragraph{Role-aware CCTO scoring.}
The aggregated CCTO score is combined with the in-category score through role-specific weights $\lambda^{n}, \lambda^{a}$. We use two scoring configurations, both selected once on a source-only validation split. A shared-role configuration uses a single coefficient $\alpha_{\mathrm{ccto}}{=}0.6$ for both roles. A role-aware configuration uses $\lambda_{n}{=}0.35$ for the normal role and $\lambda_{a}{=}0.20$ for the abnormal role, motivated by the empirical observation that the abnormal vocabulary tends to transfer more uniformly across categories than the normal vocabulary.

\paragraph{Search-time selection diagnostic.}
On frozen candidate pools regenerated under the paper's exact search configuration (provenance in \Cref{app:search_details}), we evaluated a stratified sample of 200 search survivors per direction on the actual targets (scorer-level pixel AP) and correlated the outcomes with their source-side CCTO scores; nothing feeds back into any selection. The whole-pool Spearman correlation is $+0.121$ ($p{=}0.089$) for VisA$\to$MVTec-AD and $+0.04$ ($p{=}0.62$) for MVTec-AD$\to$VisA. As a direct check, the top-scored candidate of the MVTec-AD$\to$VisA pool lands at rank 53 of 200 on target pixel AP (top 30\% of the pool), within $0.74$\,pp of the best candidate in the pool. We therefore describe CCTO as a search-time selection objective that separates transfer-brittle from transfer-stable candidates, not as a fine-grained post-hoc predictor of target performance within the surviving pool. Two effects make this within-pool readout conservative. First, a noise floor: the VisA$\to$MVTec-AD pool spans only $0.63$\,pp of target pixel AP (std $0.12$\,pp), below the pipeline's three-seed std on the same metric ($0.18$\,pp, \Cref{tab:multiseed}), so no ranking signal is resolvable there. Second, range restriction: correlations computed within CCTO-selected survivors systematically underestimate the objective's utility over the full search space. The pool spread also bounds the stakes of any within-pool pick: at most $0.36/0.74$\,pp target pixel AP on the two directions. Under strict zero-shot constraints, held-out source performance remains the only admissible selection signal; the matched-budget ablation (\Cref{tab:ablation,tab:app-same-budget-full}) provides the utility evidence for CCTO under this reading.

\section{Complete Quantitative Results}
\label{app:full_results}

\paragraph{Matched-control details.}
\label{app:matched-control-details}
Per-category $95\%$ confidence intervals on the matched-control gains in \Cref{tab:coevoad_vs_control} exclude zero on VisA$\to$MVTec-AD image AUROC ($[+0.10,+0.66]$) and pixel F1 ($[+0.06,+0.47]$), but include zero on MVTec-AD$\to$VisA (dominated by \texttt{pipe\_fryum} and \texttt{pcb4}, see \Cref{tab:app-signtest}).

\paragraph{Per-category results.}
\Cref{tab:app-v2m-percat,tab:app-m2v-percat} report every target category for the two primary transfer directions, comparing the matched control (class-name prompts, no evolved rules) against CoEvoAD. The breakdown shows the gain is concentrated in a subset of categories and is not uniform: in the VisA$\to$MVTec-AD direction, image AUROC improves on 8/15 categories and decreases on 2; in the MVTec-AD$\to$VisA direction it improves on 7/12 and decreases on 5.

\begin{table*}[!t]
  \centering
  \caption{Per-category breakdown for the MVTec direction (VisA$\to$MVTec-AD).
  C: strict class-name control. E: CoEvoAD. $\Delta=$E$-$C (pp).
  Image AUROC / pixel AP / pixel F1. Values are single-seed (seed~111);
  3-seed means per routed category are reported in \Cref{tab:app-route-audit-v2m},
  and per-category three-seed mean$\pm$std in \Cref{tab:app-percat-seedstd}.}
  \label{tab:app-v2m-percat}
  \footnotesize
  \setlength{\tabcolsep}{3.5pt}
  \begin{tabular}{l rrr rrr rrr}
    \toprule
    & \multicolumn{3}{c}{AUROC\textsubscript{img}} & \multicolumn{3}{c}{AP\textsubscript{px}} & \multicolumn{3}{c}{F1\textsubscript{px}} \\
    \cmidrule(lr){2-4}\cmidrule(lr){5-7}\cmidrule(lr){8-10}
    Category & C & E & $\Delta$ & C & E & $\Delta$ & C & E & $\Delta$ \\
    \midrule
    bottle & 94.76 & 94.84 & +0.08 & 67.19 & 66.97 & $-$0.22 & 63.09 & 62.92 & $-$0.17 \\
    cable & 82.27 & 83.36 & +1.09 & 12.49 & 12.01 & $-$0.48 & 20.09 & 19.98 & $-$0.11 \\
    capsule & 94.50 & 94.30 & $-$0.20 & 36.72 & 36.67 & $-$0.05 & 39.85 & 39.44 & $-$0.41 \\
    carpet & 100.00 & 100.00 & +0.00 & 82.69 & 82.98 & +0.29 & 74.93 & 75.22 & +0.29 \\
    grid & 100.00 & 100.00 & +0.00 & 41.38 & 40.97 & $-$0.41 & 42.70 & 42.75 & +0.05 \\
    hazelnut & 97.96 & 97.96 & +0.00 & 58.13 & 58.78 & +0.65 & 55.08 & 55.49 & +0.41 \\
    leather & 100.00 & 100.00 & +0.00 & 61.53 & 61.40 & $-$0.13 & 57.56 & 57.62 & +0.06 \\
    metal\_nut & 74.93 & 75.61 & +0.68 & 28.65 & 29.23 & +0.58 & 36.20 & 36.89 & +0.69 \\
    pill & 89.12 & 89.06 & $-$0.06 & 30.39 & 30.62 & +0.23 & 34.19 & 34.46 & +0.27 \\
    screw & 90.55 & 91.29 & +0.74 & 46.18 & 47.29 & +1.11 & 48.61 & 49.38 & +0.77 \\
    tile & 99.39 & 99.39 & +0.00 & 79.69 & 79.81 & +0.12 & 72.72 & 72.89 & +0.17 \\
    toothbrush & 93.06 & 94.72 & +1.66 & 20.98 & 22.44 & +1.46 & 24.10 & 25.26 & +1.16 \\
    transistor & 81.96 & 83.17 & +1.21 & 11.19 & 11.75 & +0.56 & 17.17 & 17.47 & +0.30 \\
    wood & 97.28 & 97.63 & +0.35 & 68.56 & 68.32 & $-$0.24 & 63.81 & 63.71 & $-$0.10 \\
    zipper & 99.37 & 99.50 & +0.13 & 66.46 & 66.63 & +0.17 & 65.30 & 65.89 & +0.59 \\
    \midrule
    Mean & 93.01 & 93.39 & +0.38 & 47.48 & 47.72 & +0.24 & 47.69 & 47.96 & +0.27 \\
    \bottomrule
  \end{tabular}
\end{table*}

\begin{table*}[!t]
  \centering
  \caption{Per-category breakdown for the MVTec-AD$\to$VisA direction.
  C: strict class-name control. E: CoEvoAD (MVTec-AD$\to$VisA main config).
  $\Delta=$E$-$C (pp). The MVTec-AD$\to$VisA gain is dominated by a few categories
  (\texttt{pipe\_fryum} +4.05 AP\textsubscript{px}, \texttt{pcb4}
  $-$3.27 AP\textsubscript{px}). Values are single-seed (seed~111);
  3-seed means per routed category are reported in \Cref{tab:app-route-audit-m2v}
  (\texttt{pipe\_fryum} $+1.96$, \texttt{pcb4} $-2.18$ AP\textsubscript{px}),
  and per-category three-seed mean$\pm$std in \Cref{tab:app-percat-seedstd}.}
  \label{tab:app-m2v-percat}
  \footnotesize
  \setlength{\tabcolsep}{3.5pt}
  \begin{tabular}{l rrr rrr rrr}
    \toprule
    & \multicolumn{3}{c}{AUROC\textsubscript{img}} & \multicolumn{3}{c}{AP\textsubscript{px}} & \multicolumn{3}{c}{F1\textsubscript{px}} \\
    \cmidrule(lr){2-4}\cmidrule(lr){5-7}\cmidrule(lr){8-10}
    Category & C & E & $\Delta$ & C & E & $\Delta$ & C & E & $\Delta$ \\
    \midrule
    candle & 90.02 & 90.67 & +0.65 & 33.69 & 33.53 & $-$0.16 & 44.08 & 43.55 & $-$0.53 \\
    capsules & 92.95 & 91.95 & $-$1.00 & 49.56 & 52.27 & +2.71 & 54.92 & 55.78 & +0.86 \\
    cashew & 93.84 & 93.36 & $-$0.48 & 36.35 & 37.40 & +1.05 & 42.16 & 42.68 & +0.52 \\
    chewinggum & 97.38 & 97.00 & $-$0.38 & 83.18 & 83.12 & $-$0.06 & 75.45 & 75.53 & +0.08 \\
    fryum & 90.68 & 90.92 & +0.24 & 27.95 & 27.60 & $-$0.35 & 33.19 & 32.92 & $-$0.27 \\
    macaroni1 & 91.61 & 91.27 & $-$0.34 & 26.18 & 26.42 & +0.24 & 34.80 & 35.99 & +1.19 \\
    macaroni2 & 67.43 & 68.37 & +0.94 & 3.17 & 2.75 & $-$0.42 & 9.64 & 8.56 & $-$1.08 \\
    pcb1 & 79.36 & 80.08 & +0.72 & 10.40 & 11.91 & +1.51 & 17.75 & 19.64 & +1.89 \\
    pcb2 & 76.81 & 77.81 & +1.00 & 13.03 & 14.28 & +1.25 & 21.64 & 23.13 & +1.49 \\
    pcb3 & 77.09 & 77.56 & +0.47 & 19.97 & 20.60 & +0.63 & 26.41 & 27.72 & +1.31 \\
    pcb4 & 89.87 & 92.83 & +2.96 & 28.09 & 24.82 & $-$3.27 & 33.09 & 29.72 & $-$3.37 \\
    pipe\_fryum & 97.60 & 97.28 & $-$0.32 & 34.59 & 38.64 & +4.05 & 41.41 & 45.44 & +4.03 \\
    \midrule
    Mean & 87.05 & 87.43 & +0.38 & 30.51 & 31.11 & +0.60 & 36.21 & 36.72 & +0.51 \\
    \bottomrule
  \end{tabular}
\end{table*}

\paragraph{Multi-seed stability.}
\Cref{tab:multiseed} reports the per-seed results and the mean and standard deviation over three random seeds for the two main transfer directions, following the search and transfer settings in Sec.~\ref{sec:experimental_setup}. \Cref{tab:app-percat-seedstd} breaks the same three runs down per category. The largest per-category standard deviations (0.96 image-AUROC points on \texttt{toothbrush}; 1.73 on \texttt{macaroni2}; 2.17 pixel-AP points on \texttt{pipe\_fryum}) are of the same magnitude as the worst single-run per-category drops discussed in the Limitations, so per-category readings should be interpreted against the three-seed spread rather than a single run.

\begin{table}[!t]
  \centering
  \caption{Multi-seed stability over seeds 111, 222, and 333. We report per-seed results together with mean and standard deviation under the matched protocol.}
  \label{tab:multiseed}
  \scriptsize
  \setlength{\tabcolsep}{2.5pt}
  \begin{tabular}{l cc cc}
    \toprule
    & \multicolumn{2}{c}{\textbf{VisA$\to$MVTec-AD}} & \multicolumn{2}{c}{\textbf{MVTec-AD$\to$VisA}} \\
    \cmidrule(lr){2-3} \cmidrule(lr){4-5}
    & AUROC\textsubscript{img} & AP\textsubscript{px} & AUROC\textsubscript{img} & AP\textsubscript{px} \\
    \midrule
    Seed 111 & 93.39 & 47.72 & 87.43 & 31.11 \\
    Seed 222 & 93.13 & 47.37 & 87.05 & 30.79 \\
    Seed 333 & 93.26 & 47.45 & 87.46 & 31.19 \\
    Mean$\pm$std & 93.26\,{\scriptsize$\pm$0.13} & 47.51\,{\scriptsize$\pm$0.18} & 87.31\,{\scriptsize$\pm$0.23} & 31.03\,{\scriptsize$\pm$0.21} \\
    \bottomrule
  \end{tabular}
\end{table}

\begin{table}[!t]
  \centering
  \caption{Per-category three-seed mean$\pm$std of CoEvoAD over the same three independent search runs as \Cref{tab:multiseed} (deployed configuration per direction). Std is the sample standard deviation over seeds 111/222/333.}
  \label{tab:app-percat-seedstd}
  \scriptsize
  \setlength{\tabcolsep}{5pt}
  \begin{tabular}{l cc}
    \toprule
    Category & AUROC\textsubscript{img} & AP\textsubscript{px} \\
    \midrule
    \multicolumn{3}{l}{\textit{VisA$\to$MVTec-AD}} \\
    bottle & 94.84\,{\scriptsize$\pm$0.00} & 67.03\,{\scriptsize$\pm$0.20} \\
    cable & 83.55\,{\scriptsize$\pm$0.32} & 12.09\,{\scriptsize$\pm$0.20} \\
    capsule & 94.31\,{\scriptsize$\pm$0.02} & 36.60\,{\scriptsize$\pm$0.06} \\
    carpet & 100.00\,{\scriptsize$\pm$0.00} & 82.77\,{\scriptsize$\pm$0.19} \\
    grid & 100.00\,{\scriptsize$\pm$0.00} & 41.16\,{\scriptsize$\pm$0.17} \\
    hazelnut & 97.98\,{\scriptsize$\pm$0.06} & 59.03\,{\scriptsize$\pm$0.22} \\
    leather & 100.00\,{\scriptsize$\pm$0.00} & 61.05\,{\scriptsize$\pm$0.39} \\
    metal\_nut & 75.32\,{\scriptsize$\pm$0.26} & 29.07\,{\scriptsize$\pm$0.17} \\
    pill & 89.15\,{\scriptsize$\pm$0.16} & 30.52\,{\scriptsize$\pm$0.12} \\
    screw & 90.67\,{\scriptsize$\pm$0.62} & 46.75\,{\scriptsize$\pm$0.65} \\
    tile & 99.42\,{\scriptsize$\pm$0.04} & 79.43\,{\scriptsize$\pm$0.39} \\
    toothbrush & 94.17\,{\scriptsize$\pm$0.96} & 21.63\,{\scriptsize$\pm$0.77} \\
    transistor & 82.74\,{\scriptsize$\pm$0.40} & 11.48\,{\scriptsize$\pm$0.23} \\
    wood & 97.43\,{\scriptsize$\pm$0.22} & 68.19\,{\scriptsize$\pm$0.13} \\
    zipper & 99.34\,{\scriptsize$\pm$0.21} & 65.93\,{\scriptsize$\pm$0.62} \\
    \midrule
    \multicolumn{3}{l}{\textit{MVTec-AD$\to$VisA}} \\
    candle & 90.85\,{\scriptsize$\pm$0.16} & 34.44\,{\scriptsize$\pm$0.93} \\
    capsules & 92.30\,{\scriptsize$\pm$0.76} & 52.51\,{\scriptsize$\pm$1.82} \\
    cashew & 93.50\,{\scriptsize$\pm$0.80} & 36.39\,{\scriptsize$\pm$1.31} \\
    chewinggum & 97.15\,{\scriptsize$\pm$0.14} & 83.21\,{\scriptsize$\pm$0.27} \\
    fryum & 90.89\,{\scriptsize$\pm$0.15} & 28.66\,{\scriptsize$\pm$0.93} \\
    macaroni1 & 91.22\,{\scriptsize$\pm$0.06} & 26.25\,{\scriptsize$\pm$0.34} \\
    macaroni2 & 69.41\,{\scriptsize$\pm$1.73} & 2.66\,{\scriptsize$\pm$0.11} \\
    pcb1 & 80.21\,{\scriptsize$\pm$0.11} & 12.20\,{\scriptsize$\pm$1.23} \\
    pcb2 & 77.54\,{\scriptsize$\pm$0.27} & 13.80\,{\scriptsize$\pm$0.64} \\
    pcb3 & 76.11\,{\scriptsize$\pm$1.31} & 19.79\,{\scriptsize$\pm$1.13} \\
    pcb4 & 91.23\,{\scriptsize$\pm$1.39} & 25.91\,{\scriptsize$\pm$1.00} \\
    pipe\_fryum & 97.33\,{\scriptsize$\pm$0.22} & 36.55\,{\scriptsize$\pm$2.17} \\
    \bottomrule
  \end{tabular}
\end{table}

\paragraph{External industrial transfer.}
\Cref{tab:app-external-full} reports the percentage-point deltas of CoEvoAD over the matched baseline on five external targets under both MVTec-AD and VisA source conventions. The MVTec-source route gives consistent pixel AP and F1 gains while staying image-neutral on average; the VisA-source route is image-neutral and slightly negative on pixel AP and F1, so it must not be cited as a pixel-improvement result.

\begin{table*}[!t]
  \centering
  \caption{External industrial benchmarks, all five targets, both source conventions. AUROC\textsubscript{img}: control$\to$CoEvoAD (\%). $\Delta$ columns: CoEvoAD $-$ matched strict control (pp).}
  \label{tab:app-external-full}
  \footnotesize
  \setlength{\tabcolsep}{4pt}
  \begin{tabular}{l l c r r r r}
    \toprule
    Source & Target & AUROC\textsubscript{img} C$\to$E & $\Delta$AUROC\textsubscript{px} & $\Delta$AUPRO & $\Delta$AP\textsubscript{px} & $\Delta$F1\textsubscript{px} \\
    \midrule
    MVTec & BTAD & 92.81$\to$92.68 & +0.45 & +0.47 & +0.61 & +1.56 \\
    MVTec & DAGM & 93.82$\to$93.56 & $-$0.12 & $-$0.46 & +0.81 & +0.35 \\
    MVTec & DTD & 93.77$\to$93.99 & +0.12 & +0.25 & +0.04 & +0.32 \\
    MVTec & KSDD2 & 96.04$\to$95.94 & +0.01 & +0.09 & +0.14 & +0.21 \\
    MVTec & RSDD & 97.69$\to$97.62 & +0.02 & +0.10 & +2.91 & +0.99 \\
    \midrule
    MVTec & Mean & 94.83$\to$94.76 & +0.10 & +0.09 & +0.90 & +0.69 \\
    \midrule
    VisA & BTAD & 94.25$\to$94.40 & $-$0.02 & --- & +0.04 & +0.08 \\
    VisA & DAGM & 98.17$\to$98.19 & +0.00 & --- & $-$0.39 & $-$0.32 \\
    VisA & DTD & 94.25$\to$94.25 & +0.02 & --- & $-$0.66 & $-$0.55 \\
    VisA & KSDD2 & 97.42$\to$97.44 & $-$0.01 & --- & $-$0.55 & $-$0.35 \\
    VisA & RSDD & 98.89$\to$98.86 & $-$0.01 & --- & +0.27 & $-$0.19 \\
    \midrule
    VisA & Mean & 96.60$\to$96.63 & +0.00 & --- & $-$0.26 & $-$0.27 \\
    \bottomrule
  \end{tabular}
\end{table*}

\paragraph{Texture-oriented external targets.}
We also examine two texture-oriented external targets, DAGM and DTD-Synthetic. \Cref{tab:texture_external} reports image-level AUROC. On DTD-Synthetic (standard 12-class split), CoEvoAD remains close to AnomalyCLIP but trails Bayes-PFL by about two percentage points. On DAGM (evaluated on the 7-class subset used in our protocol), CoEvoAD is the strongest among methods reproduced on the same subset; AdaCLIP and MRAD numbers are taken from their original publications under the 10-class standard split. These results are consistent with a boundary of category-keyed prompt transfer: abstract texture categories have weaker semantic overlap with the object-defect vocabulary used by the source-side prompt rules. \Cref{tab:app-texture-percat} drills down to per-class DTD-Synthetic results, showing the aggregate gap is concentrated on two categories whose names have low semantic similarity with any source object-defect class.

\begin{table}[!t]
\centering
\caption{Image-level AUROC (\%) on texture-oriented external targets. $\dagger$ denotes evaluation on our 7-class DAGM subset; unmarked DAGM numbers follow the standard 10-class reporting convention.}
\label{tab:texture_external}
\small
\setlength{\tabcolsep}{6pt}
\begin{tabular}{lcc}
\toprule
Method & DAGM & DTD-Synthetic \\
\midrule
\multicolumn{3}{l}{\emph{Same 7-class DAGM subset}} \\
AnomalyCLIP$^\dagger$ & 96.3 & 94.3 \\
Bayes-PFL$^\dagger$ & 97.35 & $\mathbf{96.28}$ \\
\textbf{CoEvoAD} & $\mathbf{98.19}$ & 94.25 \\
\midrule
\multicolumn{3}{l}{\emph{Original publication reports}} \\
WinCLIP & 87.6 & 83.9 \\
AdaCLIP & 99.1 & 95.5 \\
MRAD & 98.4 & 96.0 \\
\bottomrule
\end{tabular}
\end{table}

\begin{table}[!ht]
  \centering
  \caption{Per-class image-level AUROC (\%) on DTD-Synthetic. C: Bayes-PFL (matched VisA-source reproduction). E: CoEvoAD. $\Delta=$E$-$C.}
  \label{tab:app-texture-percat}
  \scriptsize
  \setlength{\tabcolsep}{4pt}
  \begin{tabular}{l r r r}
    \toprule
    Category & C (Bayes-PFL) & E (CoEvoAD) & $\Delta$ \\
    \midrule
    blotchy     & 94.06 & 75.75 & $-18.31$ \\
    matted1     & 88.10 & 80.95 & $-7.15$ \\
    marbled2    & 99.00 & 96.50 & $-2.50$ \\
    fibrous     & 99.94 & 98.69 & $-1.25$ \\
    mesh        & 91.15 & 92.64 & $+1.49$ \\
    perforated  & 93.00 & 96.44 & $+3.44$ \\
    stratified  & 99.81 & 99.69 & $-0.12$ \\
    woven1      & 92.48 & 93.23 & $+0.75$ \\
    woven2      & 100.00 & 100.00 & $\pm 0.00$ \\
    woven3      & 99.25 & 98.31 & $-0.94$ \\
    woven4      & 98.51 & 98.88 & $+0.37$ \\
    woven5      & 100.00 & 99.90 & $-0.10$ \\
    \midrule
    Mean        & 96.28 & 94.25 & $-2.03$ \\
    \bottomrule
  \end{tabular}
\end{table}

\section{Additional Ablations}
\label{app:additional_ablations}

\paragraph{Paired category-level significance.}
\Cref{tab:app-signtest} reports, per direction and metric, the mean per-category $\Delta$, a normal-approximation 95\% confidence interval over categories, and the improved/degraded/tied counts. The VisA$\to$MVTec-AD image-AUROC interval excludes zero; the MVTec-AD$\to$VisA image-AUROC interval includes zero, so the MVTec-AD$\to$VisA image-level gain is not distinguishable from category-level noise. This is consistent with the Limitations statement that the strongest evidence is the direction-sensitive ablation and pixel-side robustness, not a uniform lift.

\begin{table}[!ht]
  \centering
  \caption{Paired category-level mean differences with 95\% confidence intervals. $\Delta=$ CoEvoAD $-$ matched control, averaged over categories ($n{=}15$ for the VisA$\to$MVTec-AD direction, $n{=}12$ for the MVTec-AD$\to$VisA direction). CI is the normal-approximation 95\% interval of the per-category $\Delta$. Differences of $\pm$0.01 from \Cref{tab:coevoad_vs_control} reflect different rounding paths (aggregate vs per-category).}
  \label{tab:app-signtest}
  \scriptsize
  \setlength{\tabcolsep}{3.5pt}
  \begin{tabular}{l l r l rrr}
    \toprule
    Dir. & Metric & Mean $\Delta$ & 95\% CI & \#Up & \#Down & \#Tie \\
    \midrule
    \multirow{3}{*}{MVTec} & AUROC\textsubscript{img} & +0.38 & [+0.10,\,+0.66] & 8 & 2 & 5 \\
                            & AP\textsubscript{px} & +0.24 & [$-$0.03,\,+0.51] & 9 & 6 & 0 \\
                            & F1\textsubscript{px} & +0.27 & [+0.06,\,+0.47] & 11 & 4 & 0 \\
    \midrule
    \multirow{3}{*}{VisA}  & AUROC\textsubscript{img} & +0.37 & [$-$0.19,\,+0.93] & 7 & 5 & 0 \\
                            & AP\textsubscript{px} & +0.60 & [$-$0.38,\,+1.58] & 7 & 5 & 0 \\
                            & F1\textsubscript{px} & +0.51 & [$-$0.47,\,+1.49] & 8 & 4 & 0 \\
    \bottomrule
  \end{tabular}
\end{table}

\paragraph{Test-time transfer policy.}
\label{app:routing}
\Cref{tab:routing} reports the test-time transfer ablation, separating nearest-source rule transfer (Sem) from template transfer (Tmpl). R1 and R3 coincide because semantic transfer covers every target category (\Cref{tab:app-route-audit-v2m,tab:app-route-audit-m2v}), so the template fallback never fires once rule transfer is enabled. The MVTec-AD$\to$VisA rows share one frozen rule set from an earlier search run, which keeps the four routes matched within the table; under the same transfer policy the deployed rule set of \Cref{tab:coevoad_vs_control} reaches 87.43 image AUROC and 31.11 pixel AP, slightly above the R3 row here, while the rule-free control row R0 matches \Cref{tab:coevoad_vs_control} exactly.

\begin{table}[!htbp]
  \centering
  \caption{Test-time transfer ablation. R0 is the class-name control; Sem/Tmpl denote nearest-source/template transfer.}
  \label{tab:routing}
  \scriptsize
  \setlength{\tabcolsep}{2pt}
  \resizebox{\columnwidth}{!}{%
  \begin{tabular}{l cc rr rr}
    \toprule
    & & & \multicolumn{2}{c}{\textbf{MVTec-AD}} & \multicolumn{2}{c}{\textbf{VisA}} \\
    \cmidrule(lr){4-5} \cmidrule(lr){6-7}
    Route & Sem & Tmpl & AUROC\textsubscript{img} & AP\textsubscript{px} & AUROC\textsubscript{img} & AP\textsubscript{px} \\
    \midrule
    R0 (control) & \xmark & \xmark & 93.01 & 47.48 & 87.05 & 30.51 \\
    R1 (sem-only) & \cmark & \xmark & \textbf{93.39} & \textbf{47.72} & \textbf{87.21} & 30.85 \\
    R2 (tmpl-only) & \xmark & \cmark & 93.13 & 47.49 & 87.11 & \textbf{31.51} \\
    R3 (both) & \cmark & \cmark & \textbf{93.39} & \textbf{47.72} & \textbf{87.21} & 30.85 \\
    \bottomrule
  \end{tabular}}
\end{table}

\paragraph{Per-class routing audit.}
\Cref{tab:app-route-audit-v2m,tab:app-route-audit-m2v} report, for every target category, the routed source donor, its semantic similarity, and the per-class deltas of CoEvoAD over the class-name control (3-seed means; the per-category breakdowns in \Cref{tab:app-v2m-percat,tab:app-m2v-percat} are single-seed). Routing is deterministic and seed-stable: all 27 target classes keep the same donor across the three seeds. No mapping fails catastrophically; the worst per-class image delta is $-0.98$\,pp (\texttt{pcb3}). On MVTec-AD$\to$VisA, 8 of 12 classes route to the same donor (\texttt{pill}) with nearly identical similarities ($0.756$--$0.826$), yet their pixel-AP deltas span $-2.18$ to $+1.96$\,pp: identical routing with opposite outcomes, pointing to rule-content $\times$ class interaction rather than donor selection.

\begin{table}[!htbp]
  \centering
  \caption{Per-class routing audit, VisA$\to$MVTec-AD: routed donor, semantic similarity, and deltas vs.\ the class-name control (pp, 3-seed means; single-seed values in \Cref{tab:app-v2m-percat}, policy tags in \Cref{tab:app-routing-v2m}). The control is a single run; seed variance is method-side (cf.\ \Cref{tab:multiseed}).}
  \label{tab:app-route-audit-v2m}
  \scriptsize
  \setlength{\tabcolsep}{3pt}
  \resizebox{\columnwidth}{!}{%
  \begin{tabular}{l l c rrr}
    \toprule
    Category & Donor & Sim & $\Delta$AUROC\textsubscript{img} & $\Delta$AP\textsubscript{px} & $\Delta$F1\textsubscript{px} \\
    \midrule
    bottle & candle & 0.810 & +0.08 & $-$0.16 & $-$0.09 \\
    cable & macaroni1 & 0.758 & +1.28 & $-$0.40 & +0.01 \\
    capsule & capsules & 0.903 & $-$0.19 & $-$0.12 & $-$0.34 \\
    carpet & fryum & 0.799 & +0.00 & +0.08 & +0.07 \\
    grid & fryum & 0.803 & +0.00 & $-$0.22 & +0.11 \\
    hazelnut & cashew & 0.848 & +0.02 & +0.90 & +0.56 \\
    leather & fryum & 0.749 & +0.00 & $-$0.48 & $-$0.20 \\
    metal\_nut & cashew & 0.794 & +0.39 & +0.42 & +0.46 \\
    pill & macaroni1 & 0.826 & +0.03 & +0.13 & +0.13 \\
    screw & macaroni1 & 0.759 & +0.12 & +0.57 & +0.30 \\
    tile & fryum & 0.791 & +0.03 & $-$0.26 & $-$0.17 \\
    toothbrush & candle & 0.786 & +1.11 & +0.65 & +0.48 \\
    transistor & fryum & 0.736 & +0.78 & +0.29 & +0.11 \\
    wood & fryum & 0.786 & +0.15 & $-$0.37 & $-$0.19 \\
    zipper & macaroni1 & 0.816 & $-$0.03 & $-$0.53 & $-$0.14 \\
    \midrule
    Mean & & & +0.25 & +0.03 & +0.07 \\
    \bottomrule
  \end{tabular}}
\end{table}

\begin{table}[!htbp]
  \centering
  \caption{Per-class routing audit, MVTec-AD$\to$VisA (pp, 3-seed means; single-seed values in \Cref{tab:app-m2v-percat}, policy tags in \Cref{tab:app-routing-m2v}).}
  \label{tab:app-route-audit-m2v}
  \scriptsize
  \setlength{\tabcolsep}{3pt}
  \resizebox{\columnwidth}{!}{%
  \begin{tabular}{l l c rrr}
    \toprule
    Category & Donor & Sim & $\Delta$AUROC\textsubscript{img} & $\Delta$AP\textsubscript{px} & $\Delta$F1\textsubscript{px} \\
    \midrule
    candle & bottle & 0.810 & +0.83 & +0.75 & $-$0.06 \\
    capsules & capsule & 0.903 & $-$0.65 & +2.95 & +1.66 \\
    cashew & hazelnut & 0.848 & $-$0.34 & +0.04 & $-$0.15 \\
    chewinggum & toothbrush & 0.716 & $-$0.23 & +0.03 & +0.12 \\
    fryum & pill & 0.819 & +0.21 & +0.71 & +0.39 \\
    macaroni1 & pill & 0.826 & $-$0.39 & +0.07 & +0.88 \\
    macaroni2 & pill & 0.806 & +1.98 & $-$0.51 & $-$1.41 \\
    pcb1 & pill & 0.770 & +0.85 & +1.80 & +2.04 \\
    pcb2 & pill & 0.775 & +0.73 & +0.77 & +0.94 \\
    pcb3 & pill & 0.756 & $-$0.98 & $-$0.18 & +0.23 \\
    pcb4 & pill & 0.764 & +1.36 & $-$2.18 & $-$2.40 \\
    pipe\_fryum & pill & 0.770 & $-$0.27 & +1.96 & +2.23 \\
    \midrule
    Mean & & & +0.26 & +0.52 & +0.37 \\
    \bottomrule
  \end{tabular}}
\end{table}

\paragraph{Donor-forced counterfactual matrix.}
To bound the cost of routing errors, we force every target class onto every source donor (exact-rule reinstantiation with only the class-name slot replaced) and evaluate each forced assignment with the frozen scorer (test-only inference, PRO skipped). \Cref{tab:app-cf-aggregate} reports dataset-level deltas vs.\ the frozen seed-111 route re-materialized on the same surface; \Cref{tab:app-cf-envelope-v2m,tab:app-cf-envelope-m2v} report the per-class envelope. All rows are comparable only within this family; absolute values are not comparable with \Cref{tab:coevoad_vs_control} or \Cref{tab:app-v2m-percat,tab:app-m2v-percat}. The oracle assignment selects the best donor per class using target labels: it is not zero-shot and serves only as an upper reference. Within the evaluated bank the cost of routing errors is bounded: the actual route beats the uniform-random expectation on all metrics (by $+0.1$--$0.25$\,pp), and even the adversarial worst assignment, which itself requires target labels to construct, costs at most $0.54/2.26$\,pp image AUROC and $0.95/2.47$\,pp pixel AP on the two directions.

\begin{table}[!htbp]
  \centering
  \caption{Donor-forced counterfactual aggregates ($\Delta$ vs.\ the actual route, pp; negative = worse). Anchor row: absolute scores of the actual route under this surface. Oracle uses target labels (not zero-shot; reference only).}
  \label{tab:app-cf-aggregate}
  \scriptsize
  \setlength{\tabcolsep}{3pt}
  \resizebox{\columnwidth}{!}{%
  \begin{tabular}{l c c}
    \toprule
    & MVTec & VisA \\
    Forced route & \multicolumn{2}{c}{AUROC\textsubscript{img} / AP\textsubscript{px} / F1\textsubscript{px}} \\
    \midrule
    Actual route (anchor, absolute) & 93.39 / 47.72 / 47.96 & 87.43 / 31.11 / 36.72 \\
    Random donor (expectation) & $-$0.21 / $-$0.25 / $-$0.28 & $-$0.25 / $-$0.11 / $-$0.02 \\
    Worst donor (adversarial) & $-$0.54 / $-$0.95 / $-$0.88 & $-$2.26 / $-$2.47 / $-$2.35 \\
    Oracle donor (not zero-shot) & +0.14 / +0.48 / +0.35 & +1.47 / +2.29 / +2.29 \\
    \bottomrule
  \end{tabular}}
\end{table}

\begin{table*}[!htbp]
  \centering
  \caption{Per-class donor-forced counterfactual envelope, VisA$\to$MVTec-AD (absolute scores under the forced-route surface; in-family comparison only). Act.: actual route; Rand.: uniform-random expectation over donors; Worst/Orac.: per-class minimum/maximum over the donor bank (oracle uses target labels; not zero-shot).}
  \label{tab:app-cf-envelope-v2m}
  \scriptsize
  \setlength{\tabcolsep}{3pt}
  \begin{tabular}{l rrrr rrrr rrrr}
    \toprule
    & \multicolumn{4}{c}{AUROC\textsubscript{img}} & \multicolumn{4}{c}{AP\textsubscript{px}} & \multicolumn{4}{c}{F1\textsubscript{px}} \\
    \cmidrule(lr){2-5}\cmidrule(lr){6-9}\cmidrule(lr){10-13}
    Category & Act. & Rand. & Worst & Orac. & Act. & Rand. & Worst & Orac. & Act. & Rand. & Worst & Orac. \\
    \midrule
    bottle & 94.84 & 94.92 & 94.76 & 95.16 & 66.97 & 67.88 & 66.79 & 68.46 & 62.92 & 63.62 & 62.78 & 64.09 \\
    cable & 83.36 & 83.09 & 82.38 & 83.85 & 12.01 & 12.61 & 12.01 & 13.19 & 19.98 & 20.36 & 19.98 & 20.99 \\
    capsule & 94.30 & 94.46 & 94.14 & 94.81 & 36.67 & 36.89 & 36.36 & 37.36 & 39.44 & 39.84 & 39.34 & 40.40 \\
    carpet & 100.00 & 100.00 & 100.00 & 100.00 & 82.98 & 82.76 & 82.50 & 83.03 & 75.22 & 74.98 & 74.76 & 75.22 \\
    grid & 100.00 & 99.99 & 99.92 & 100.00 & 40.97 & 41.22 & 40.80 & 41.63 & 42.75 & 42.70 & 42.11 & 43.04 \\
    hazelnut & 97.96 & 98.00 & 97.79 & 98.14 & 58.78 & 58.09 & 56.95 & 59.72 & 55.49 & 54.84 & 53.79 & 56.25 \\
    leather & 100.00 & 100.00 & 100.00 & 100.00 & 61.40 & 61.27 & 60.44 & 62.11 & 57.62 & 57.40 & 56.68 & 58.12 \\
    metal\_nut & 75.61 & 75.19 & 74.78 & 75.61 & 29.23 & 28.97 & 28.23 & 29.89 & 36.89 & 36.50 & 36.11 & 36.89 \\
    pill & 89.06 & 89.19 & 88.76 & 89.50 & 30.62 & 30.45 & 30.21 & 30.63 & 34.46 & 34.14 & 33.86 & 34.46 \\
    screw & 91.29 & 90.41 & 89.65 & 91.29 & 47.29 & 45.55 & 43.51 & 47.29 & 49.38 & 47.93 & 46.02 & 49.38 \\
    tile & 99.39 & 99.41 & 99.39 & 99.46 & 79.81 & 79.47 & 78.99 & 79.81 & 72.89 & 72.61 & 72.28 & 72.89 \\
    toothbrush & 94.72 & 93.84 & 93.06 & 94.72 & 22.44 & 21.31 & 20.27 & 22.44 & 25.26 & 24.24 & 23.35 & 25.26 \\
    transistor & 83.17 & 82.40 & 82.08 & 83.17 & 11.75 & 11.36 & 11.22 & 11.75 & 17.47 & 17.23 & 17.02 & 17.47 \\
    wood & 97.63 & 97.46 & 97.02 & 97.72 & 68.32 & 68.35 & 67.91 & 69.14 & 63.71 & 63.71 & 63.49 & 64.28 \\
    zipper & 99.50 & 99.28 & 99.05 & 99.50 & 66.63 & 65.96 & 65.48 & 66.63 & 65.89 & 65.14 & 64.67 & 65.89 \\
    \midrule
    Mean & 93.39 & 93.18 & 92.85 & 93.53 & 47.72 & 47.48 & 46.78 & 48.21 & 47.96 & 47.68 & 47.08 & 48.31 \\
    \bottomrule
  \end{tabular}
\end{table*}

\begin{table*}[!htbp]
  \centering
  \caption{Per-class donor-forced counterfactual envelope, MVTec-AD$\to$VisA (same protocol and reading as \Cref{tab:app-cf-envelope-v2m}).}
  \label{tab:app-cf-envelope-m2v}
  \scriptsize
  \setlength{\tabcolsep}{3pt}
  \begin{tabular}{l rrrr rrrr rrrr}
    \toprule
    & \multicolumn{4}{c}{AUROC\textsubscript{img}} & \multicolumn{4}{c}{AP\textsubscript{px}} & \multicolumn{4}{c}{F1\textsubscript{px}} \\
    \cmidrule(lr){2-5}\cmidrule(lr){6-9}\cmidrule(lr){10-13}
    Category & Act. & Rand. & Worst & Orac. & Act. & Rand. & Worst & Orac. & Act. & Rand. & Worst & Orac. \\
    \midrule
    candle & 90.67 & 90.63 & 89.79 & 91.75 & 33.53 & 33.96 & 31.99 & 36.12 & 43.55 & 43.94 & 42.61 & 45.79 \\
    capsules & 91.95 & 92.81 & 91.60 & 94.03 & 52.27 & 51.45 & 44.31 & 55.32 & 55.78 & 56.38 & 49.35 & 59.93 \\
    cashew & 93.36 & 93.33 & 92.46 & 94.20 & 37.40 & 37.41 & 35.47 & 40.90 & 42.68 & 43.01 & 41.50 & 45.55 \\
    chewinggum & 97.00 & 97.15 & 96.88 & 97.48 & 83.12 & 82.99 & 81.19 & 83.82 & 75.53 & 75.43 & 74.35 & 76.00 \\
    fryum & 90.92 & 91.01 & 90.48 & 92.36 & 27.60 & 28.45 & 27.06 & 30.50 & 32.92 & 33.64 & 32.47 & 35.17 \\
    macaroni1 & 91.27 & 91.50 & 90.66 & 92.94 & 26.42 & 25.93 & 24.67 & 26.82 & 35.99 & 35.10 & 34.14 & 36.09 \\
    macaroni2 & 68.37 & 68.53 & 62.47 & 73.43 & 2.75 & 3.30 & 2.00 & 4.34 & 8.56 & 9.61 & 6.68 & 12.04 \\
    pcb1 & 80.08 & 79.48 & 74.81 & 81.43 & 11.91 & 12.18 & 9.15 & 15.89 & 19.64 & 19.62 & 16.70 & 22.91 \\
    pcb2 & 77.81 & 77.28 & 74.82 & 80.26 & 14.28 & 13.52 & 11.80 & 15.65 & 23.13 & 22.35 & 20.80 & 25.01 \\
    pcb3 & 77.56 & 76.33 & 73.62 & 77.76 & 20.60 & 20.17 & 18.10 & 22.53 & 27.72 & 26.88 & 25.22 & 28.59 \\
    pcb4 & 92.83 & 90.87 & 87.75 & 92.96 & 24.82 & 27.03 & 24.82 & 30.26 & 29.72 & 31.93 & 29.12 & 35.57 \\
    pipe\_fryum & 97.28 & 97.23 & 96.64 & 98.12 & 38.64 & 35.68 & 33.09 & 38.64 & 45.44 & 42.47 & 39.49 & 45.44 \\
    \midrule
    Mean & 87.43 & 87.18 & 85.17 & 88.89 & 31.11 & 31.01 & 28.64 & 33.40 & 36.72 & 36.70 & 34.37 & 39.01 \\
    \bottomrule
  \end{tabular}
\end{table*}

\section{Prompt-Rule Interpretability Analysis}
\label{app:prompt_analysis}

\paragraph{Selected prompt rules per source category.}
\Cref{tab:app-rules-v2m,tab:app-rules-m2v} list the evolved prompt-rule pair selected per source category by the locked rule search. ``\texttt{X}'' denotes the learned soft-context slot of the prompt bank (the prompt to CLIP is hybrid: a learned latent context plus the interpretable rule string), so these are interpretable \emph{rule strings}, not a fully interpretable system. The default column is the class-name baseline the search starts from; the evolved column is the saved, reloadable selection. The search is mutation-only (no recombination).

\begin{table*}[!tp]
  \centering
  \caption{Evolved prompt rules per VisA source category, for the VisA$\to$MVTec-AD direction.
  Default $=$ \texttt{X normal/abnormal \{cat\}}.}
  \label{tab:app-rules-v2m}
  \scriptsize
  \setlength{\tabcolsep}{4pt}
  \begin{tabular}{l p{0.42\textwidth} p{0.42\textwidth}}
    \toprule
    Src cat. & Evolved normal rule & Evolved abnormal rule \\
    \midrule
    candle & X clean intact candle & X visible anomaly on object \\
    capsules & X a photo of a capsules & X plain damaged capsules \\
    cashew & X a photo of healthy cashew & X faulty flawed cashew \\
    chewinggum & X typical intact with chewinggum & X anomalous abnormal textural chewinggum \\
    fryum & X a view of pristine good fryum & X item: defect in fryum \\
    macaroni1 & X a picture of macaroni1 & X defective visible textural macaroni1 \\
    macaroni2 & X perfect macaroni2 & X object: in macaroni2 \\
    pcb1 & X typical flawless perfect pcb1 & X item: irregular pcb1 \\
    pcb2 & X flawless fine pcb2 & X structural defect like a hole or cut \\
    pcb3 & X clean regular pcb3 & X a view of compromised imperfect pcb3 \\
    pcb4 & X a photo of a pcb4 & X faulty flawed deteriorated pcb4 \\
    pipe\_fryum & X object: flawless healthy pipe\_fryum & X object: faulty pipe\_fryum \\
    \bottomrule
  \end{tabular}
\end{table*}

\begin{table*}[!tp]
  \centering
  \caption{Evolved prompt rules per MVTec-AD source category, for the MVTec-AD$\to$VisA direction. Default $=$ \texttt{X normal/abnormal \{cat\}}.}
  \label{tab:app-rules-m2v}
  \scriptsize
  \setlength{\tabcolsep}{4pt}
  \begin{tabular}{l p{0.42\textwidth} p{0.42\textwidth}}
    \toprule
    Src cat. & Evolved normal rule & Evolved abnormal rule \\
    \midrule
    bottle & X normal bottle & X irregular visible surface bottle \\
    cable & X typical cable & X an anomalous cable sample \\
    capsule & X a standard capsule product & X a deformed structure capsule sample \\
    carpet & X pristine surface carpet & X faulty blemished carpet \\
    grid & X regular standard grid & X faulty flawed abnormal grid \\
    hazelnut & X hazelnut & X abnormal faulty flawed hazelnut \\
    leather & X perfect leather & X faulty damaged region with structural leather \\
    metal\_nut & X perfect metal\_nut & X broken faulty compromised metal\_nut \\
    pill & X pill & X item: irregular deteriorated pill \\
    screw & X clean surface screw & X broken shape with missing parts \\
    tile & X the normal tile & X plain flawed tile \\
    toothbrush & X a photo of a healthy toothbrush & X faulty toothbrush \\
    transistor & X clean transistor & X defect in defective irregular transistor \\
    wood & X standard good wood & X defective irregular wood \\
    zipper & X regular zipper & X defective bad deformed zipper \\
    \bottomrule
  \end{tabular}
\end{table*}

\paragraph{Target transfer-policy map.}
\Cref{tab:app-routing-v2m,tab:app-routing-m2v} give the transfer-policy decision for every target category. Each target category is mapped to its nearest source category by text-embedding cosine similarity, and the target category name is the only target-side input.

\begin{table}[!tp]
  \centering
  \caption{VisA$\to$MVTec-AD transfer policy: MVTec-AD target $\to$ VisA source. Sim $=$ cosine similarity; Tag $=$ policy tag.}
  \label{tab:app-routing-v2m}
  \footnotesize
  \setlength{\tabcolsep}{2pt}
  \renewcommand{\arraystretch}{1.02}
  \begin{tabular}{@{}l l r l@{}}
    \toprule
    Target & Nearest source & Sim & Route tag \\
    \midrule
    bottle & candle & 0.810 & semantic (abn.\ fallback) \\
    cable & macaroni1 & 0.758 & semantic transfer \\
    capsule & capsules & 0.903 & semantic transfer \\
    carpet & fryum & 0.799 & semantic transfer \\
    grid & fryum & 0.804 & semantic transfer \\
    hazelnut & cashew & 0.848 & semantic transfer \\
    leather & fryum & 0.749 & semantic transfer \\
    metal\_nut & cashew & 0.794 & semantic transfer \\
    pill & macaroni1 & 0.826 & semantic transfer \\
    screw & macaroni1 & 0.759 & semantic transfer \\
    tile & fryum & 0.791 & semantic transfer \\
    toothbrush & candle & 0.786 & semantic (abn.\ fallback) \\
    transistor & fryum & 0.736 & semantic transfer \\
    wood & fryum & 0.786 & semantic transfer \\
    zipper & macaroni1 & 0.816 & semantic transfer \\
    \bottomrule
  \end{tabular}
\end{table}

\begin{table}[!tp]
  \centering
  \caption{MVTec-AD$\to$VisA transfer policy: VisA target $\to$ MVTec-AD source.}
  \label{tab:app-routing-m2v}
  \footnotesize
  \setlength{\tabcolsep}{2pt}
  \renewcommand{\arraystretch}{1.02}
  \begin{tabular}{@{}l l r l@{}}
    \toprule
    Target & Nearest source & Sim & Route tag \\
    \midrule
    candle & bottle & 0.810 & semantic transfer \\
    capsules & capsule & 0.903 & semantic transfer \\
    cashew & hazelnut & 0.848 & semantic transfer \\
    chewinggum & toothbrush & 0.716 & semantic transfer \\
    fryum & pill & 0.819 & semantic transfer \\
    macaroni1 & pill & 0.826 & semantic transfer \\
    macaroni2 & pill & 0.806 & semantic transfer \\
    pcb1 & pill & 0.770 & semantic transfer \\
    pcb2 & pill & 0.775 & semantic transfer \\
    pcb3 & pill & 0.756 & semantic transfer \\
    pcb4 & pill & 0.764 & semantic transfer \\
    pipe\_fryum & pill & 0.770 & semantic transfer \\
    \bottomrule
  \end{tabular}
\end{table}

\begin{figure}[!ht]
  \centering
  \includegraphics[width=\columnwidth]{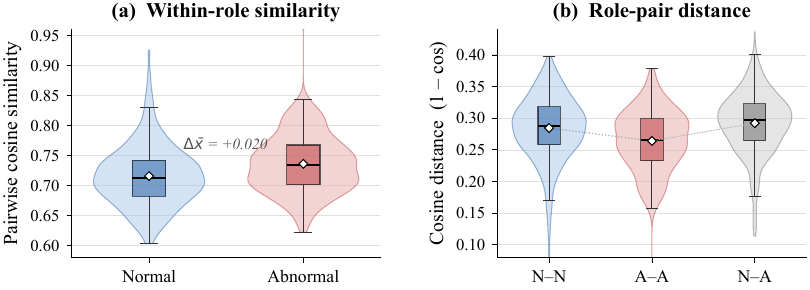}
  \caption{Analysis of selected rule embeddings. We report within-role cosine similarity and cross-role distance for the text embeddings of selected normal and abnormal rules.}
  \label{Fig:prompt_embed}
\end{figure}

\paragraph{Prompt embedding analysis.}
\Cref{Fig:prompt_embed} reports the embedding-level view of the selected rules: pairwise within-role cosine similarity for the normal and abnormal populations, and the distances between rule pairs within and across roles.

\section{Additional Qualitative Results and Failure Cases}
\label{app:qualitative}

\begin{figure*}[!ht]
  \centering
  \includegraphics[width=\textwidth]{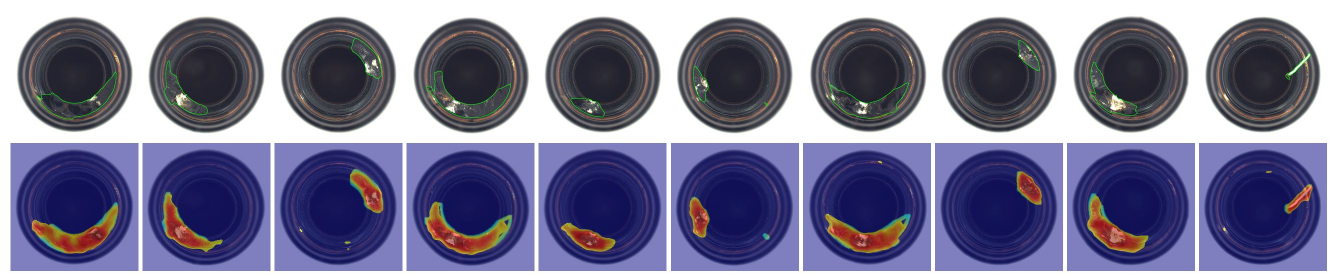}
  \caption{Qualitative localization on MVTec-AD \textit{bottle}. Top: input with ground-truth contours; bottom: CoEvoAD anomaly map.}
  \label{Fig:app-qual-mvtec-bottle}
\end{figure*}

\begin{figure*}[!ht]
  \centering
  \includegraphics[width=\textwidth]{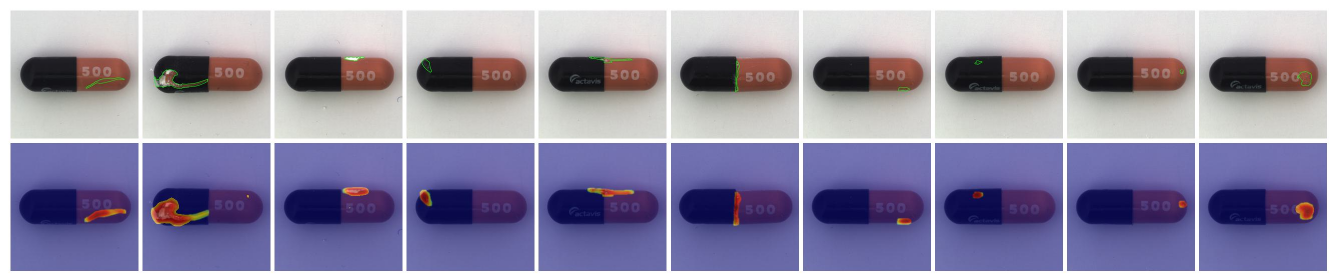}
  \caption{Qualitative localization on MVTec-AD \textit{capsule}.}
  \label{Fig:app-qual-mvtec-capsule}
\end{figure*}

\begin{figure*}[!ht]
  \centering
  \includegraphics[width=\textwidth]{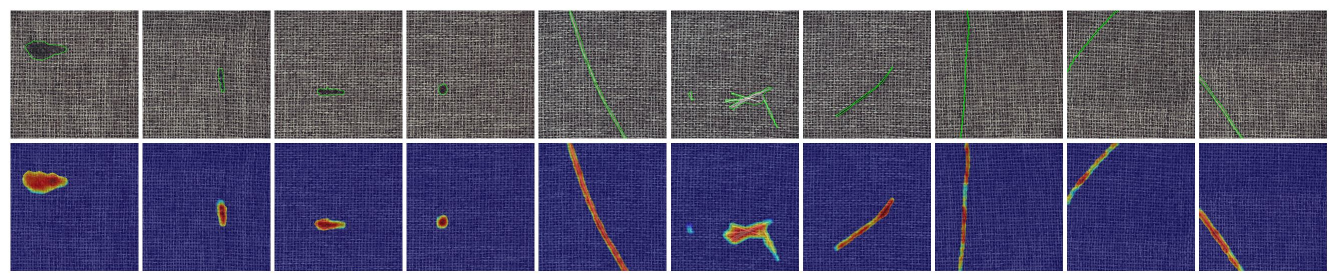}
  \caption{Qualitative localization on MVTec-AD \textit{carpet}.}
  \label{Fig:app-qual-mvtec-carpet}
\end{figure*}

\begin{figure*}[!ht]
  \centering
  \includegraphics[width=\textwidth]{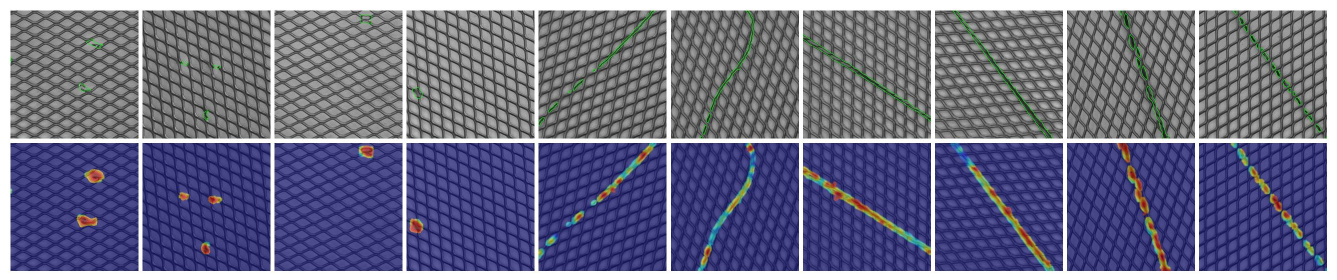}
  \caption{Qualitative localization on MVTec-AD \textit{grid}.}
  \label{Fig:app-qual-mvtec-grid}
\end{figure*}

\begin{figure*}[!ht]
  \centering
  \includegraphics[width=\textwidth]{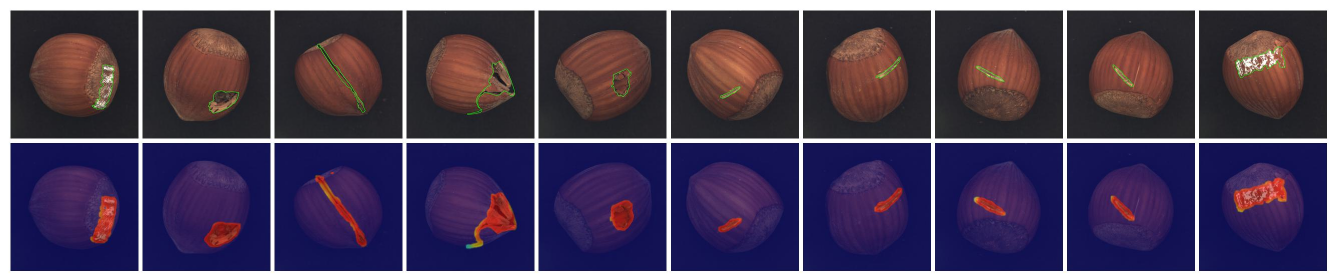}
  \caption{Qualitative localization on MVTec-AD \textit{hazelnut}.}
  \label{Fig:app-qual-mvtec-hazelnut}
\end{figure*}

\begin{figure*}[!ht]
  \centering
  \includegraphics[width=\textwidth]{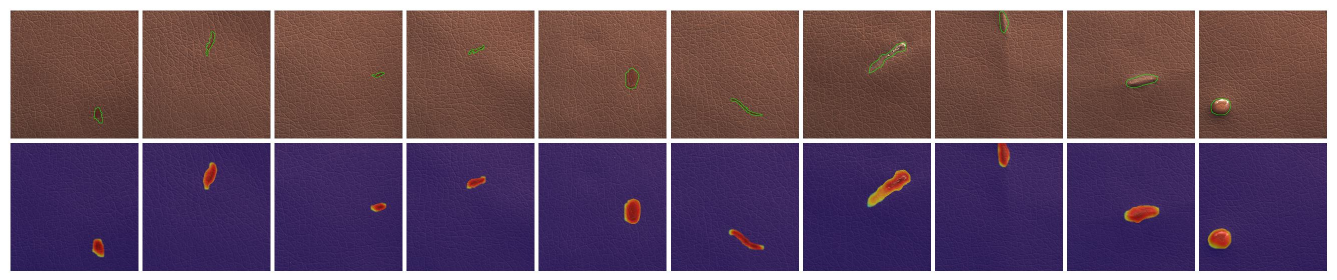}
  \caption{Qualitative localization on MVTec-AD \textit{leather}.}
  \label{Fig:app-qual-mvtec-leather}
\end{figure*}

\begin{figure*}[!ht]
  \centering
  \includegraphics[width=\textwidth]{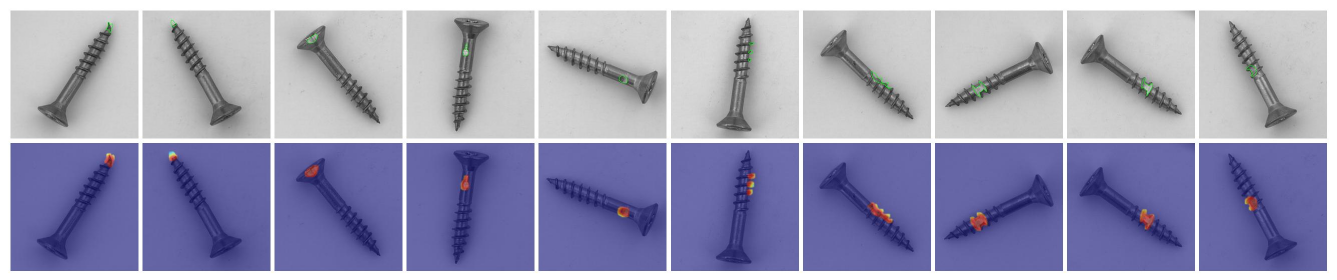}
  \caption{Qualitative localization on MVTec-AD \textit{screw}.}
  \label{Fig:app-qual-mvtec-screw}
\end{figure*}

\begin{figure*}[!ht]
  \centering
  \includegraphics[width=\textwidth]{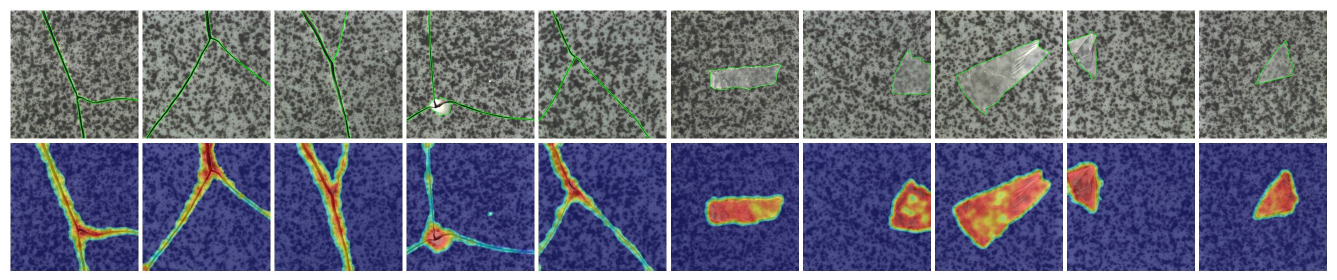}
  \caption{Qualitative localization on MVTec-AD \textit{tile}.}
  \label{Fig:app-qual-mvtec-tile}
\end{figure*}

\begin{figure*}[!ht]
  \centering
  \includegraphics[width=\textwidth]{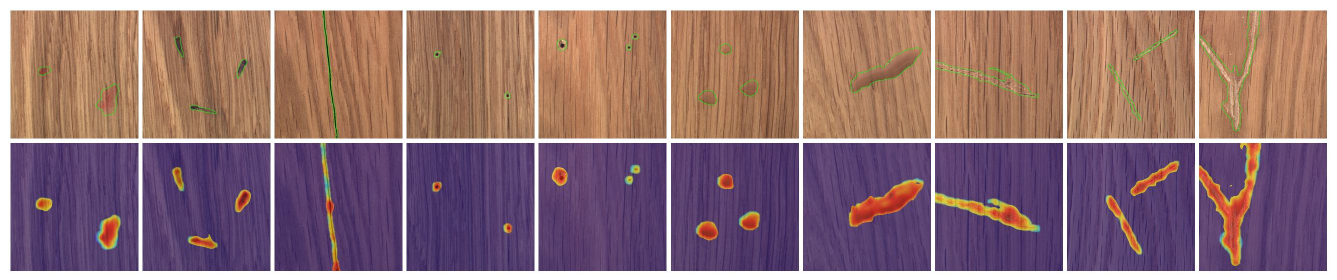}
  \caption{Qualitative localization on MVTec-AD \textit{wood}.}
  \label{Fig:app-qual-mvtec-wood}
\end{figure*}

\begin{figure*}[!ht]
  \centering
  \includegraphics[width=\textwidth]{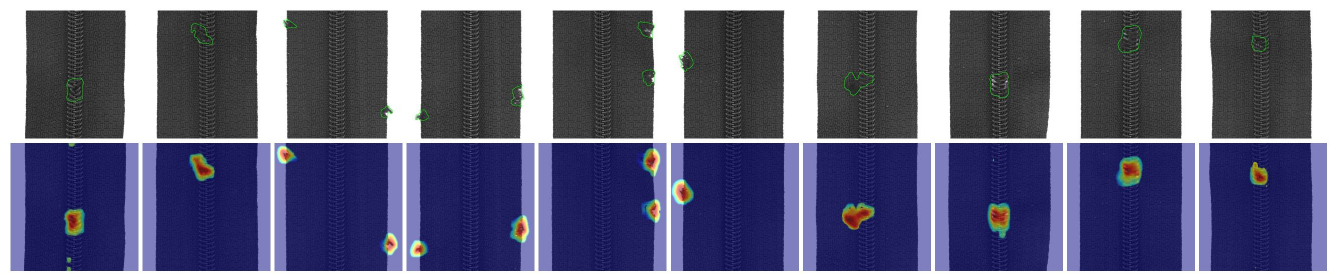}
  \caption{Qualitative localization on MVTec-AD \textit{zipper}.}
  \label{Fig:app-qual-mvtec-zipper}
\end{figure*}

\begin{figure*}[!ht]
  \centering
  \includegraphics[width=\textwidth]{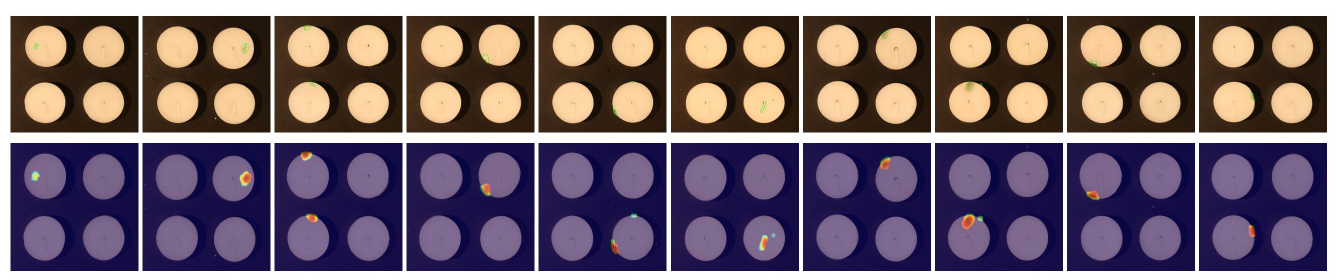}
  \caption{Qualitative localization on VisA \textit{candle}.}
  \label{Fig:app-qual-visa-candle}
\end{figure*}

\begin{figure*}[!ht]
  \centering
  \includegraphics[width=\textwidth]{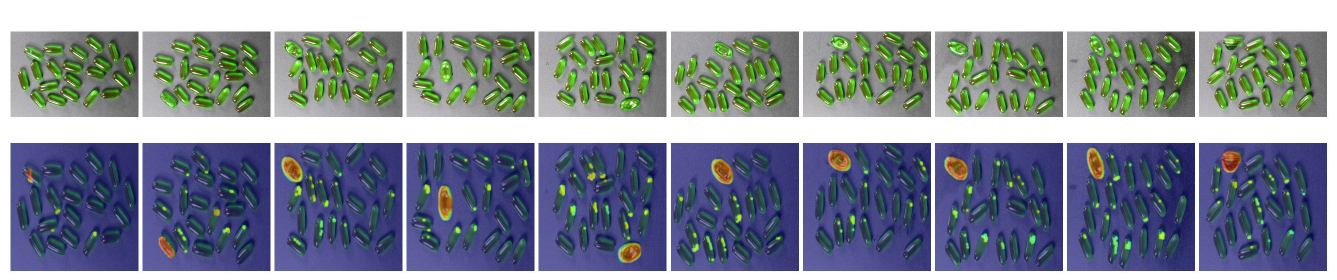}
  \caption{Qualitative localization on VisA \textit{capsules}.}
  \label{Fig:app-qual-visa-capsules}
\end{figure*}

\begin{figure*}[!ht]
  \centering
  \includegraphics[width=\textwidth]{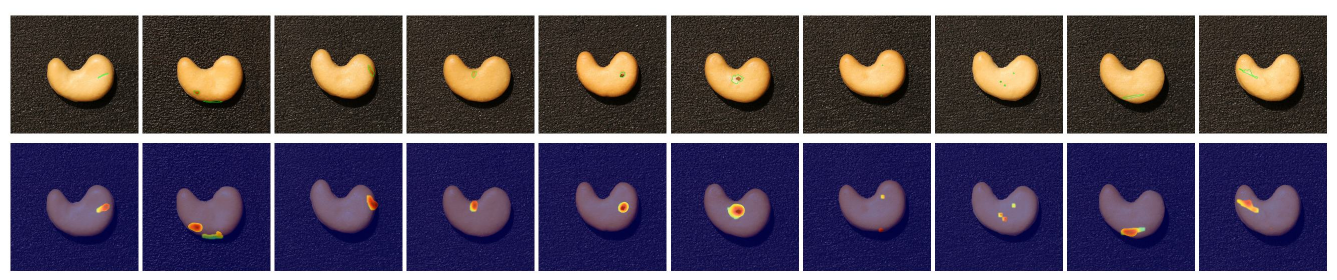}
  \caption{Qualitative localization on VisA \textit{cashew}.}
  \label{Fig:app-qual-visa-cashew}
\end{figure*}

\begin{figure*}[!ht]
  \centering
  \includegraphics[width=\textwidth]{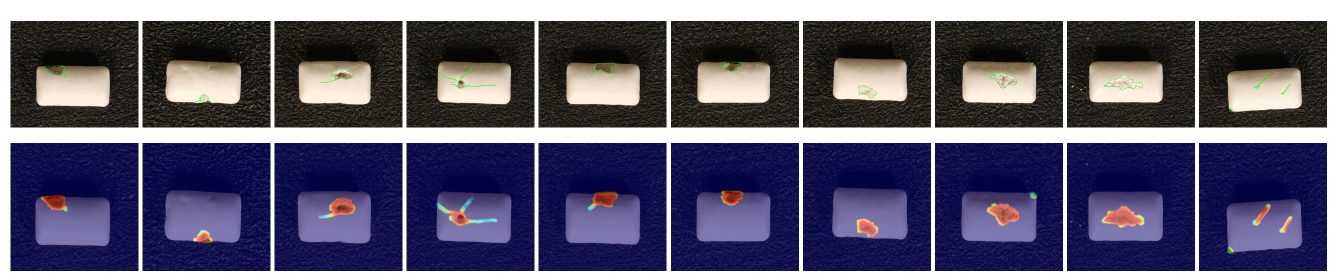}
  \caption{Qualitative localization on VisA \textit{chewinggum}.}
  \label{Fig:app-qual-visa-chewinggum}
\end{figure*}

\begin{figure*}[!ht]
  \centering
  \includegraphics[width=\textwidth]{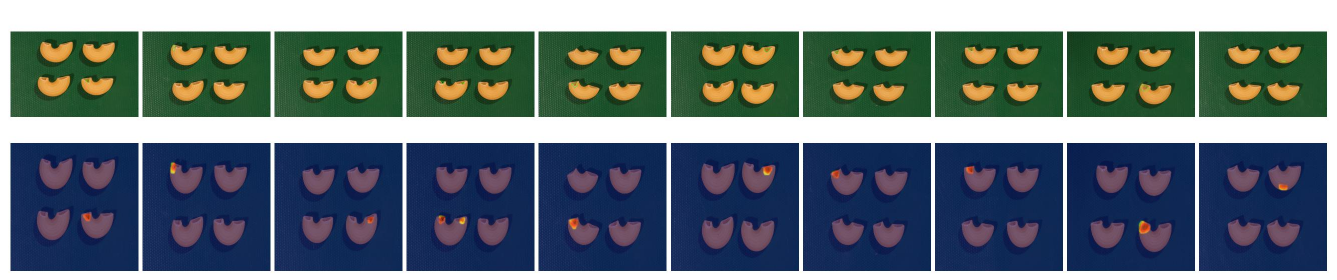}
  \caption{Qualitative localization on VisA \textit{macaroni1}.}
  \label{Fig:app-qual-visa-macaroni1}
\end{figure*}

\begin{figure*}[!ht]
  \centering
  \includegraphics[width=\textwidth]{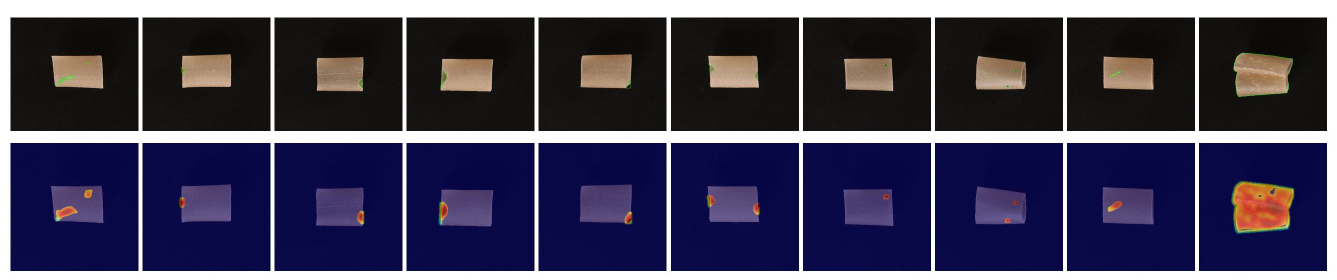}
  \caption{Qualitative localization on VisA \textit{pipe\_fryum}.}
  \label{Fig:app-qual-visa-pipe-fryum}
\end{figure*}

\begin{figure*}[!ht]
  \centering
  \includegraphics[width=\textwidth]{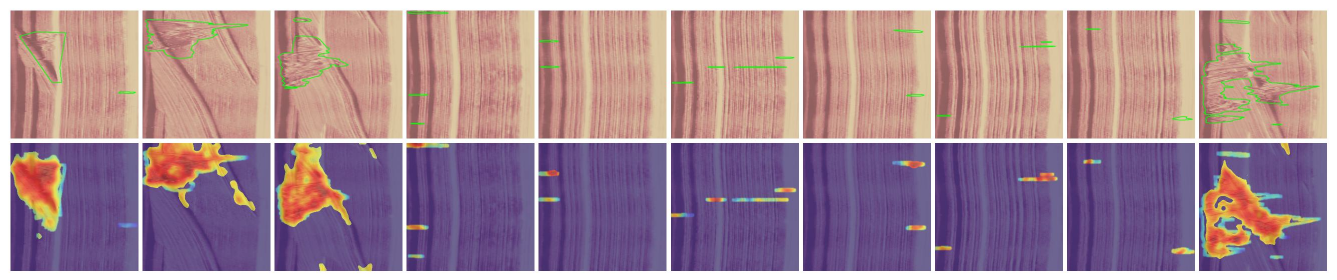}
  \caption{Qualitative localization on BTAD \textit{wood}.}
  \label{Fig:app-qual-btad-wood}
\end{figure*}

\begin{figure*}[!ht]
  \centering
  \includegraphics[width=\textwidth]{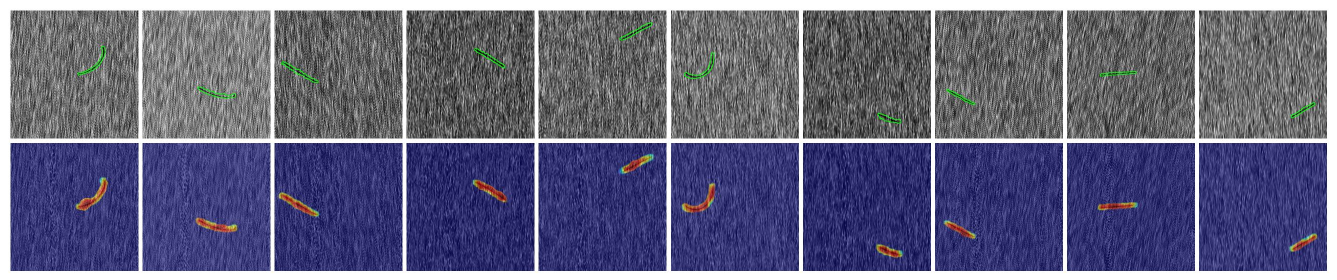}
  \caption{Qualitative localization on DAGM \textit{fabric2}.}
  \label{Fig:app-qual-dagm-fabric2}
\end{figure*}

\begin{figure*}[!ht]
  \centering
  \includegraphics[width=\textwidth]{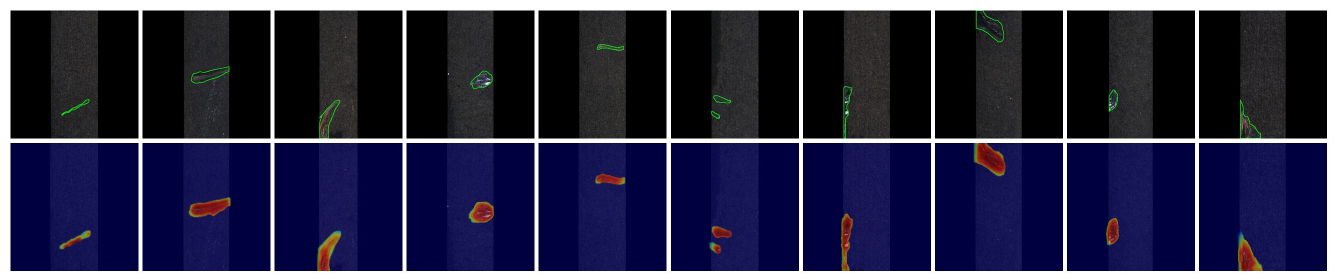}
  \caption{Qualitative localization on KSDD2 \textit{metal}.}
  \label{Fig:app-qual-ksdd2-metal}
\end{figure*}

\begin{figure*}[!ht]
  \centering
  \includegraphics[width=\textwidth]{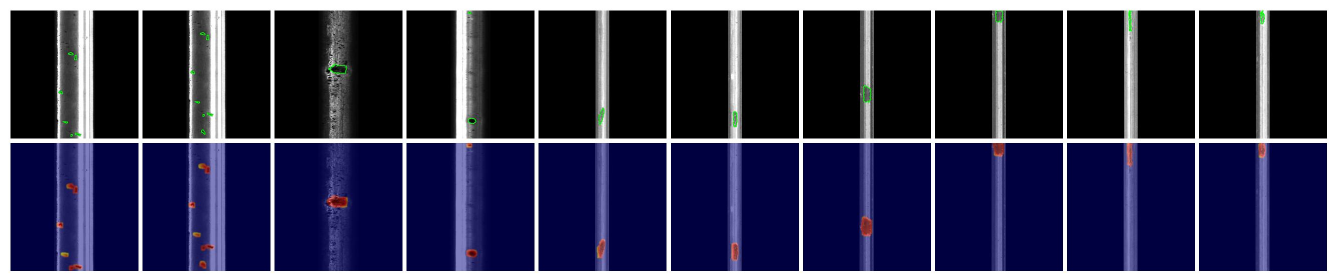}
  \caption{Qualitative localization on RSDD \textit{metal15}.}
  \label{Fig:app-qual-rsdd-metal15}
\end{figure*}

\paragraph{Qualitative localization.}
The main paper (\Cref{Fig:qualitative_maps}) compares anomaly maps for a subset of MVTec-AD categories. For completeness, \crefrange{Fig:app-qual-mvtec-bottle}{Fig:app-qual-rsdd-metal15} report the full per-category anomaly localization produced by CoEvoAD on the two primary cross-dataset directions (MVTec, VisA) and the four external industrial targets (BTAD, DAGM, KSDD2, RSDD). In every figure, the top row shows input images with ground-truth defect contours overlaid in green, and the bottom row shows the CoEvoAD anomaly map (warmer colors indicate higher anomaly scores). All maps are produced by the locked rule set.

\paragraph{MRAD reproduction for \Cref{Fig:qualitative_maps}.}
The MRAD row in \Cref{Fig:qualitative_maps} is generated with the official implementation and the released source-only checkpoints (training and test datasets disjoint), using the same variant and protocol as the quantitative comparison in \Cref{tab:main-image,tab:main-pixel} (ViT-L/14@336px at resolution 518, cross-direction memory bank). In our environment the released checkpoints reproduce the published cross-domain metrics within $0.5$\,pp on all eight image- and pixel-level metrics across the two directions. The anomaly maps use the same test images and the same per-image min--max normalization as the other learned-prompt rows of the figure.

\paragraph{Failure modes.}
Three failure modes are visible across the per-category breakdowns; together they mark the boundaries of the method and the scope of the claim.

\emph{(i) Texture-only targets with weak object-defect overlap.}
On DTD-Synthetic, two categories (\texttt{blotchy}, \texttt{matted1}) drive the entire aggregate gap (\Cref{tab:app-texture-percat}). Their category names are abstract texture descriptors with no clear semantic counterpart in the source object-defect vocabulary used during rule search, so neither nearest-source rule transfer nor the template fallback finds a strong source anchor.

\emph{(ii) PCB-style targets with high intra-category variability.}
On MVTec-AD$\to$VisA the pixel-AP delta is dominated by \texttt{pipe\_fryum} ($+4.05$\,pp) and \texttt{pcb4} ($-3.27$\,pp), and the per-category mean-difference interval includes zero (\Cref{tab:app-signtest}). The two-direction asymmetry (VisA$\to$MVTec-AD image-AUROC interval excludes zero; MVTec-AD$\to$VisA image-AUROC interval does not) is consistent with this category-level instability and is discussed in the Limitations.

\emph{(iii) Image-saturated categories.}
On targets where the matched control already exceeds 99 image AUROC (\eg, several MVTec textures), CoEvoAD has no remaining headroom and reports $\pm 0.00$\,pp. These rows are not failures of the search but reflect the ceiling under the current backbone.

\clearpage

\end{document}